\documentclass[9pt,twocolumn,twoside]{pnas-new}

\setboolean{displaywatermark}{false}

\templatetype{pnasresearcharticle} 

\usepackage{bm, mathrsfs, graphics,float,amssymb,amsmath,subeqnarray,setspace,graphicx,amsthm,epstopdf,color}
\usepackage{enumitem}
\usepackage{soul}

\usepackage[utf8]{inputenc}
\usepackage[T1]{fontenc}

\usepackage[font=small]{caption}
\usepackage[export]{adjustbox}
\usepackage{subcaption}

\usepackage{tikz}
\tikzset{
  initial/.tip = {},
  arrowstart/.style={<-initial}
}
\usepackage{textpos}

\newcommand\ssb{\mathrm{SS}_{\textnormal{b}}}
\newcommand\ssw{\mathrm{SS}_{\textnormal{w}}}
\newtheorem{theorem}{Theorem}
\newtheorem{proposition}[theorem]{Proposition}

\renewcommand{\mathbf}{\bm}
\renewcommand{\epsilon}{\varepsilon}

\begin{document}

\title{A Law of Data Separation in Deep Learning}


\author[a,b]{Hangfeng He} 
\author[c,1]{Weijie J.~Su}

\affil[a]{Department of Computer Science, University of Rochester, Rochester, NY 14627}
\affil[b]{Goergen Institute for Data Science, University of Rochester, Rochester, NY 14627}
\affil[c]{Department of Statistics and Data Science, University of Pennsylvania, Philadelphia, PA 19104}

\leadauthor{He and Su} 

\significancestatement{The practice of deep learning has long been shrouded in mystery, leading many to believe that the inner workings of these black-box models are chaotic during training. In this paper, we challenge this belief by presenting a simple and approximate law that deep neural networks follow when processing data in the intermediate layers. This empirical law is observed in a class of modern network architectures for vision tasks, and its emergence is shown to bring important benefits for the trained models. The significance of this law is that it allows for a new perspective that provides useful insights into the practice of deep learning.}

\authorcontributions{H.H and W.J.S.~designed research, performed research, analyzed data, and wrote the paper.}
\authordeclaration{The authors declare no competing interest.}
\correspondingauthor{\textsuperscript{1}To whom correspondence should be addressed. E-mail: suw@wharton.upenn.edu.}

\keywords{deep learning $|$ data separation $|$ constant geometric rate $|$ intermediate layers} 

\begin{abstract}
While deep learning has enabled significant advances in many areas of science, its black-box nature hinders architecture design for future artificial intelligence applications and interpretation for high-stakes decision makings. We addressed this issue by studying the fundamental question of how deep neural networks process data in the intermediate layers. Our finding is a simple and quantitative law that governs how deep neural networks separate data according to class membership throughout all layers for classification. This law shows that each layer improves data separation at a constant geometric rate, and its emergence is observed in a collection of network architectures and datasets during training. This law offers practical guidelines for designing architectures, improving model robustness and out-of-sample performance, as well as interpreting the predictions.
\end{abstract}

\dates{This manuscript was compiled on \today}

\maketitle
\thispagestyle{firststyle}
\ifthenelse{\boolean{shortarticle}}{\ifthenelse{\boolean{singlecolumn}}{\abscontentformatted}{\abscontent}}{}



\begin{figure*}[!htp]
	\centering
	\includegraphics[scale=1.0]{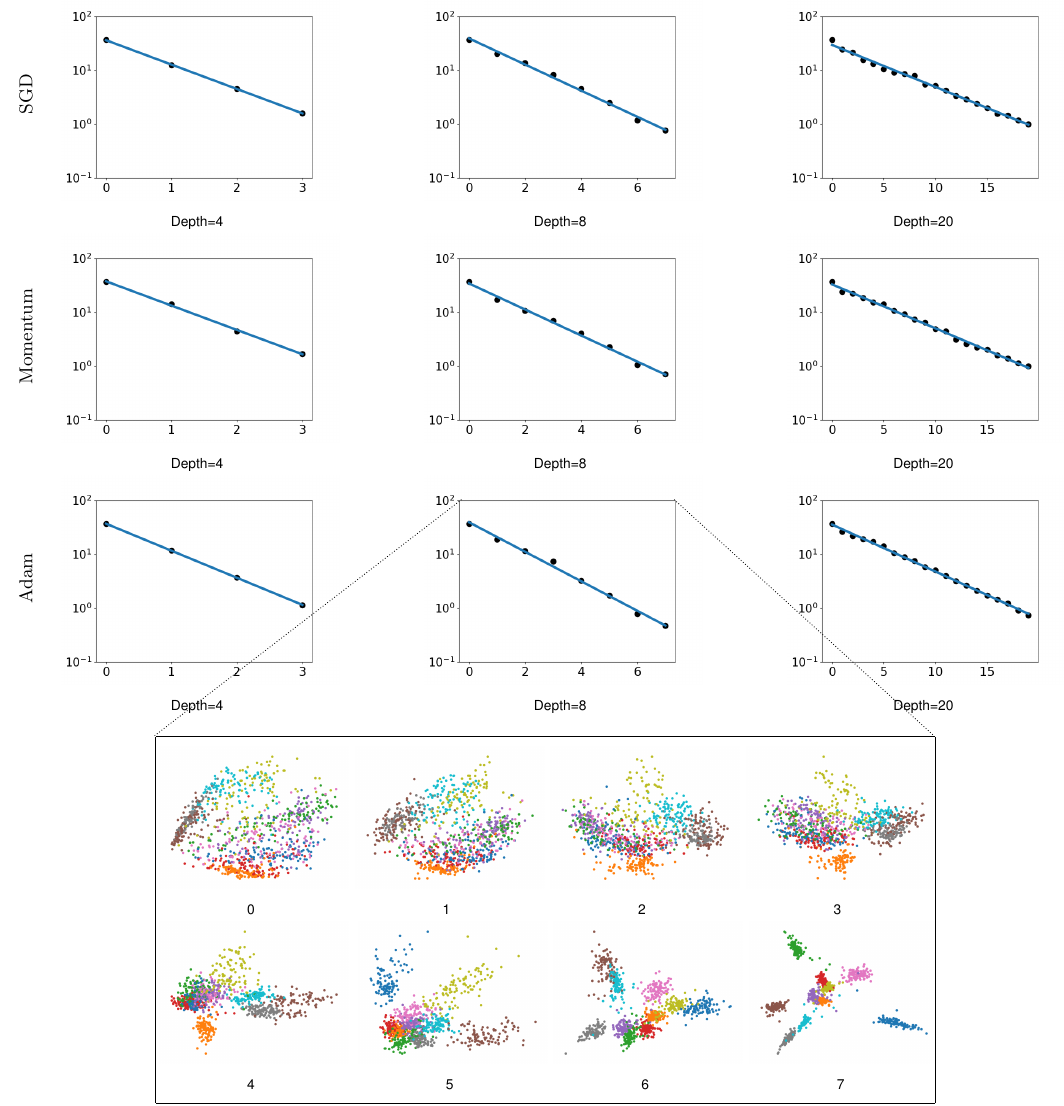}
    \caption{Illustration of the law of equi-separation in feedforward neural networks with ReLU activations trained on the Fashion-MNIST dataset. The three rows correspond to three different training methods, stochastic gradient descent (SGD), SGD with momentum, and Adam~\citep{kingma2015adam}. Throughout the paper, the $x$ axis represents the layer index and the $y$ axis represents the separation fuzziness defined in Eq.~\ref{eq:trace}, unless otherwise specified. The Pearson correlation coefficients, by row first, are $-1.000$, $-0.998$, $-0.997$, $-0.999$, $-0.998$, $-0.998$, $-1.000$, $-0.997$, and $-0.999$. The last two rows show the intermediate data points on the plane of the first two principal components for the 8-layer network trained using Adam. Layer 0 corresponds to the raw input. More details can be found in SI Appendix.
		}
	\label{fig:optimization}
\end{figure*}

\dropcap{D}eep learning methodologies have achieved remarkable success across a wide range of data-intensive tasks in image recognition, biological research, and scientific computing~\cite{krizhevsky2012imagenet,lecun2015deep, silver2016mastering,fawzi2022discovering}. In contrast to other machine learning techniques~\cite{hastie2009elements}, however, the practice of deep learning relies heavily on a plethora of heuristics and tricks that are not well justified. This situation often makes deep learning-based approaches ungrounded for some applications or necessitates the need for huge computational resources for exhaustive search, making it difficult to fully realize the potential of this set of methodologies~\cite{hutson2018has}.

This unfortunate situation is in part owing to a lack of understanding of how the prediction depends on the intermediate layers of deep neural networks~\cite{bartlett2020benign,lu2018beyond,fang2021exploring}. In particular, little is known about how the data of different classes (e.g., images of cats and dogs) in classification problems are gradually separated from the bottom layers to the top layers in modern architectures such as AlexNet~\cite{krizhevsky2012imagenet} and residual neural networks~\cite{he2016deep}. Any knowledge about data separation, especially quantitative characterization, would offer useful principles and insights for designing network architectures, training processes, and model interpretation.

The main finding of this paper is a quantitative delineation of the data separation process throughout all layers of deep neural networks. As an illustration, Fig.~\ref{fig:optimization} plots a certain value that measures how well the data are separated according to their class membership at each layer for feedforward neural networks trained on the Fashion-MNIST dataset \cite{xiao2017fashion}. This value in the logarithmic scale decays, in a distinct manner, linearly in the number of layers the data have passed through. The Pearson correlation coefficients between the logarithm of this value and the layer index range from $-0.997$ and $-1$ in Fig.~\ref{fig:optimization}. 

The measure shown in Fig.~\ref{fig:optimization} is canonical for measuring data separation in classification problems. Let $x_{ki}$ denote an intermediate output of a neural network on the $i$th point of Class $k$ for $1 \le i \le n_k$, $\bar x_{k}$ denote the sample mean of Class $k$, and $\bar x$ denote the mean of all $n := n_1 + \cdots + n_K$ data points. We define the between-class sum of squares and the within-class sum of squares as
\[
\begin{aligned}
&\ssb := \frac1{n} \sum_{k=1}^K n_k (\bar x_{k} - \bar x)(\bar x_{k} - \bar x)^\top\\
&\ssw := \frac{1}{n} \sum_{k=1}^K\sum_{i=1}^{n_k} (x_{ki} - \bar x_{k}) (x_{ki} - \bar x_{k})^\top,
\end{aligned}
\]
respectively. The former matrix represents the between-class ``signal'' for classification, whereas the latter denotes the within-class variability. Writing $\ssb^+$ for the Moore--Penrose inverse of $\ssb$ (The matrix $\ssb$ has rank at most $K-1$ and is not invertible in general since the dimension of the data is typically larger than the number of classes.), the ratio matrix $\ssw \, \ssb^+$ can be thought of as the inverse signal-to-noise ratio. We use its trace
\begin{equation}\label{eq:trace}
D := \mathrm{Tr}(\ssw \, \ssb^+)
\end{equation}
\parshape=0
to measure how well the data are separated~\cite{stevens2012applied,papyan2020prevalence}. This value, which is referred to as the separation fuzziness, is large when the data points are not concentrated to their class means or, equivalently, are not well separated, and vice versa.

\section{Main Results}
Given an $L$-layer feedforward neural network, let $D_l$ denote the separation fuzziness (Eq.~\ref{eq:trace}) of the training data passing through the first $l$ layers for $0 \le l \le L-1$.\footnote{For clarification, $D_0$ is calculated from the raw data, and $D_1$ is calculated from the data that have passed through the first layer but not the second layer.} Fig.~\ref{fig:optimization} suggests that the dynamics of $D_l$ follows the relation
\begin{equation}\label{eq:law}
D_l \doteq \rho^{l} D_0
\end{equation}
for some decay ratio $0 < \rho < 1$. Alternatively, this law implies $\log D_{l+1} - \log D_l \doteq - \log\frac1{\rho}$, showing that the neural network makes equal progress in reducing $\log D$ over each layer on the training data. Hence, we call this the {\em law of equi-separation}. This law is the first quantitative and geometric characterization of the data separation process in the intermediate layers. Indeed, it is unexpected because the intermediate output of the neural network does not exhibit any quantitative patterns, as shown by the last two rows of Fig.~\ref{fig:optimization}.

The decay ratio $\rho$ depends on the depth of the neural network, dataset, training time, and network architecture, and is also affected, to a lesser extent, by optimization methods and many other hyperparameters. For the 20-layer network trained using Adam~\cite{kingma2015adam} (the bottom-right plot of Fig.~\ref{fig:optimization}), the decay ratio $\rho$ is $0.818$. Thus, the half-life is $\frac{\ln 2}{\ln \rho^{-1}} = \frac{0.693}{\ln \rho^{-1}} = 3.45$, suggesting that this 20-layer neural network reduces the value of the separation fuzziness in every three and a half layers.

\begin{figure*}[!htp]
	\centering
	\includegraphics[scale=0.98]{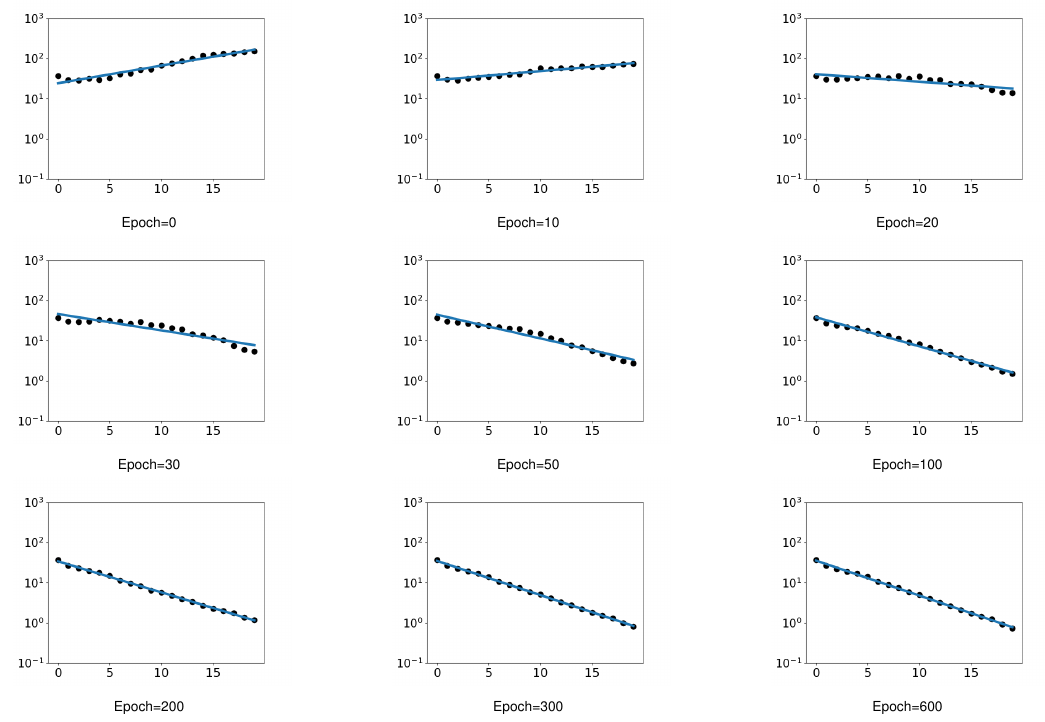}
    \caption{A 20-layer feedforward neural network trained on Fashion-MNIST. The law of equi-separation starts to emerge at epoch 100 and becomes more clear as training proceeds. As in Fig.~\ref{fig:optimization} and all the other figures in the main text, the $x$ axis represents the layer index and the $y$ axis represents the separation fuzziness.}
	\label{fig:training-epoch}
\end{figure*}

This law manifests in the terminal phase of training \citep{papyan2020prevalence}, where we continue to train the model to interpolate in-sample data. At initialization, the separation fuzziness may even increase from the bottom to top layers. During the early stages of training, the bottom layers tend to learn faster at reducing the separation fuzziness compared to the top layers 
(see Fig.~S7 in SI Appendix).
However, as training progresses, the top layers eventually catch up as the bottom layers have learned the necessary features. Finally, each layer becomes roughly equally capable of reducing the separation fuzziness multiplicatively. This dynamics of data separation during training is illustrated in Fig.~\ref{fig:training-epoch}. Neural networks in the terminal phase of training also exhibit certain symmetric geometries in the last layer such as neural collapse~\citep{papyan2020prevalence,han2021neural} and minority collapse~\cite{fang2021exploring}. However, it is worthwhile mentioning that the law of equi-separation emerges earlier than neural collapse during training (see Fig.~S1 in SI Appendix).

\begin{figure*}[!htp]
	\centering
	\includegraphics[scale=1.0]{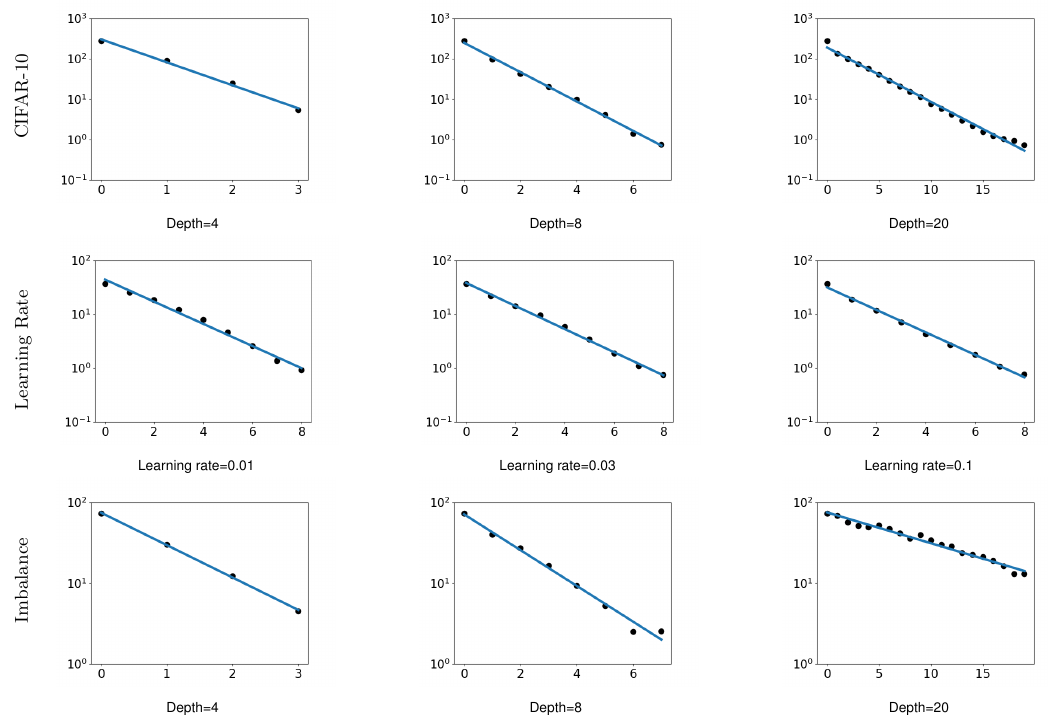}
    \caption{Illustration showing how the law of equi-separation holds in a wide range of settings. The experimental details are provided in SI Appendix.
  }
		\label{fig:elaborate}
\end{figure*}

\begin{figure}[!htp]
	\centering
	\includegraphics[scale=0.95]{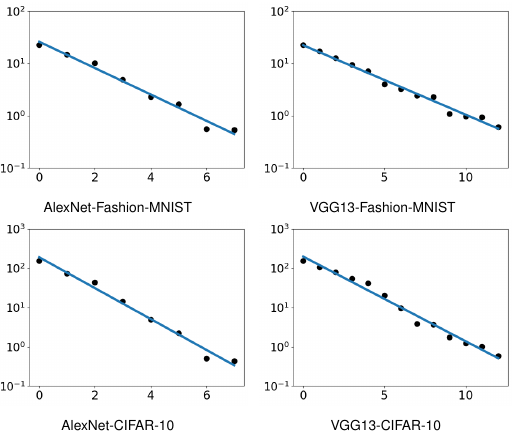}
    \caption{Illustration of the equi-separation law on convolutional neural networks. A more in-depth examination of convolutional neural networks can be found in Figs.~S2 and S3 in SI Appendix. 
  }
		\label{fig:CNNs}
\end{figure}

The pervasive law of equi-separation consistently prevails across diverse datasets, learning rates, and class imbalances, as illustrated in Fig.~\ref{fig:elaborate}. Additionally, Fig.~S6 in SI Appendix 
demonstrates its applicability in a finer, class-wise context. This law is further exemplified in contemporary network architectures for vision tasks, such as AlexNet and VGGNet~\cite{simonyan2014very}, as shown in Fig.~\ref{fig:CNNs} 
(see SI Appendix for additional convolutional neural network experiments in Fig.~S2 and Fig.~S3).
Moreover, the law manifests in residual neural networks and densely connected convolutional networks~\cite{huang2017densely} when separation fuzziness is assessed at each block, as depicted in Fig.~\ref{fig:resnet_arc} and Fig.~\ref{fig:DenseNet}, respectively. Intriguingly, this law appears slightly more pronounced in feedforward neural networks when compared with other network architectures.

The separation dynamics of neural networks have been extensively investigated in prior research studies~\cite{alain2016understanding, papyan2020traces, galanti2021role,galanti2022note, ben2022nearest}. For instance, \cite{alain2016understanding} employed linear classifiers as probes to assess the separability of intermediate outputs. In \cite{papyan2020traces}, the author scrutinized the separation capabilities of neural networks through empirical spectral analysis. More recently, \cite{galanti2021role,galanti2022note, ben2022nearest} explored the separation ability of neural networks by examining neural collapse at intermediate layers and its relationship with generalization. In particular, \cite{ben2022nearest} provided crucial experimental evidence illustrating the progression of neural collapse within the interior of neural networks, which is perhaps the most relevant work to our paper.

\section{Insights from the Law}
Next, we demonstrate the benefits of the law of equi-separation and the perspective of data separation in regard to the three fundamental aspects of deep learning practice, namely, architecture design, training, and interpretation.

\subsection{Network architecture} The law of equi-separation can inform the design principles for network architecture. First, the law implies that neural networks should be deep for good performance, thereby reflecting the hierarchical nature of deep learning \cite{bengio2009learning, lecun2015deep, schmidhuber2015deep, hihi1995hierarchical}. This is because, as shown by Eq.~\ref{eq:law}, all layers reduce the separation fuzziness $D_0$ of the raw input to $\rho^{L-1} D_0$ at the last layer. When $L$ is small---say, $L =2$ or 3---the ratio $\rho^{L-1}$ would generally not be small and, therefore, the neural network is unlikely to separate the data well. In the literature, the fundamental role of depth is recognized by analyzing the loss functions~\cite{bengio2009learning, he2016deep, glorot2010understanding, yarotsky2017error, simonyan2014very}, and our law of equi-separation offers a new perspective on network depth.

\begin{figure*}[!htp]
	\centering
	\includegraphics[scale=1.0]{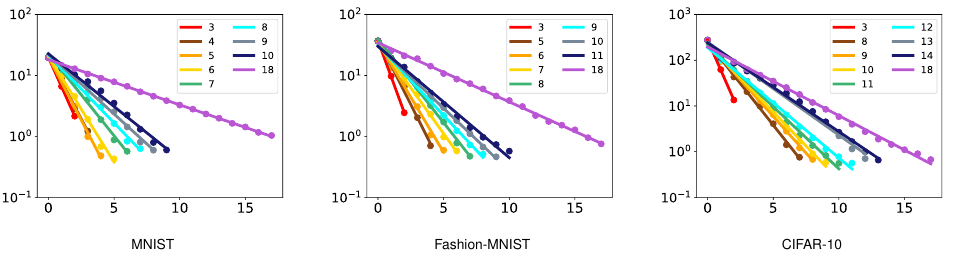}
    \caption{Illustration of the law of equi-separation with varying network depths. The legend shows how depths are color coded. The depth that yields the lowest separation fuzziness depends on the dataset complexity.}
		\label{fig:layer-number-dataset}
\end{figure*}

\begin{figure*}[!htp]
	\centering
	\includegraphics[scale=1.0]{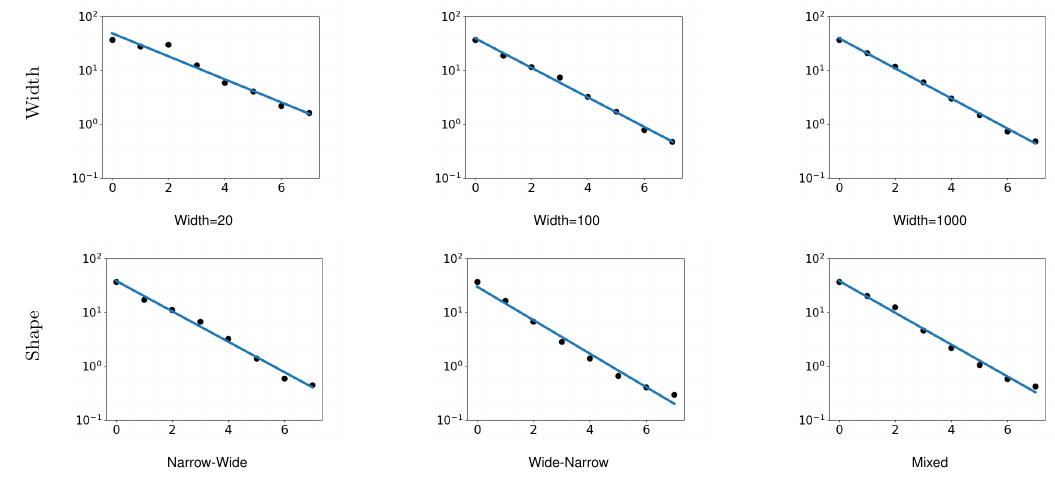}
    \caption{Illustration of the law of equi-separation with different widths and shapes of the neural networks. For example, ``Narrow-Wide'' means that the bottom layers are narrower than the top layers.} 
			\label{fig:architectures}
\end{figure*}

For completeness, the deeper the better is not necessarily true because as the depth $L$ increase, $\rho$ might become larger. Moreover, a very large depth would render the optimization problem challenging. As shown in Fig.~\ref{fig:optimization}, the 20-layer neural networks have higher values of the final separation fuzziness than those of their 8-layer counterparts. The Fashion-MNIST dataset is not very complex and, therefore, an 8-layer network will suffice to classify the data well. Adding additional layers, therefore, would only give rise to optimization burdens. Indeed, Fig.~\ref{fig:layer-number-dataset} illustrates the law of equi-separation with varying network depths, showing that different datasets correspond to different optimal depths. Therefore, as a practical guideline, the choice of depth should consider the complexity of the applications. For instance, this perspective of data separation shows that the optimal depths for MNIST~\cite{lecun1998mnist}, Fashion-MNIST, and CIFAR-10~\cite{krizhevsky2009learning} are 6, 10, and 12, respectively. This is consistent with the increasing complexity from MNIST to Fashion-MNIST to CIFAR-10~\cite{he2015convolutional, srivastava2015training, he2016deep}.

Moreover, the law of equi-separation indicates that depth bears a more significant connection to training performance than the ``shape'' of neural networks. In the first row of Fig.~\ref{fig:architectures}, the network does not separate the data well when the width of each layer is merely 20. This is because the network has to be wide enough to pass the useful features of the data for classification. However, when the network is wide enough, the network performance may be saturated even if more neurons are added to each layer~\cite{howard2017mobilenets}. In addition, the second row of Fig.~\ref{fig:architectures} shows that it is better to start with relatively wide layers at the bottom layers followed by narrow top layers. These observations suggest that very wide neural networks in general should not be recommended as they consume more time for optimization. Apart from encoding purposes, the width should be set to about the same across all layers or be made larger for the bottom layers~\cite{he2015convolutional,tan2019efficientnet}.

\subsection{Training} 
The emergence of the law of equi-separation during training is indicative of good model performance. For example, its manifestation can improve the robustness to model shifts. Indeed, the data separation ability of the network, as measured by $\frac{D_{L-1}}{D_0}$, becomes least sensitive to perturbations in network weights when the law manifests. To see this point, let
\[
R:= \frac{D_{L-1}}{D_{L-2}} \times \frac{D_{L-2}}{D_{L-3}} \times \cdots \times \frac{D_1}{D_0} = \frac{D_{L-1}}{D_0}
\]
denote the reduction ratio of the separation fuzziness made by the entire network. Assuming that a perturbation of the network weights leads to a change of $\epsilon$ in the ratio for each layer, the perturbed reduction ratio becomes
\begin{equation}\label{eq:homo}
\begin{aligned}
R_{\epsilon} &:= \left(\frac{D_{L-1}}{D_{L-2}} + \epsilon \right) \left(\frac{D_{L-2}}{D_{L-3}} + \epsilon\right) \cdots \left(\frac{D_1}{D_0} + \epsilon\right) \\
&= R + R\left(\frac{D_{L-2}}{D_{L-1}} + \frac{D_{L-3}}{D_{L-2}} +  \cdots + \frac{D_0}{D_1}\right)\epsilon + O(\epsilon^2).
\end{aligned}
\end{equation}
Applying the inequality of arithmetic and geometric means, the perturbation term $R\left(\frac{D_{L-2}}{D_{L-1}} + \frac{D_{L-3}}{D_{L-2}} +  \cdots + \frac{D_0}{D_1}\right)\epsilon$ is minimized in absolute value when $\frac{D_{L-1}}{D_{L-2}} = \frac{D_{L-2}}{D_{L-3}} = \cdots =\frac{D_1}{D_0}$. This result is summarized in the following proposition:

\begin{proposition}\label{prop}
For small $\epsilon$, $|R_{\epsilon} - R|$ is minimized when 
\[
\frac{D_{L-1}}{D_{L-2}} = \frac{D_{L-2}}{D_{L-3}} = \cdots =\frac{D_1}{D_0}.
\]
\end{proposition}

\begin{figure*}[!htp]
	\centering
	\includegraphics[scale=1.0]{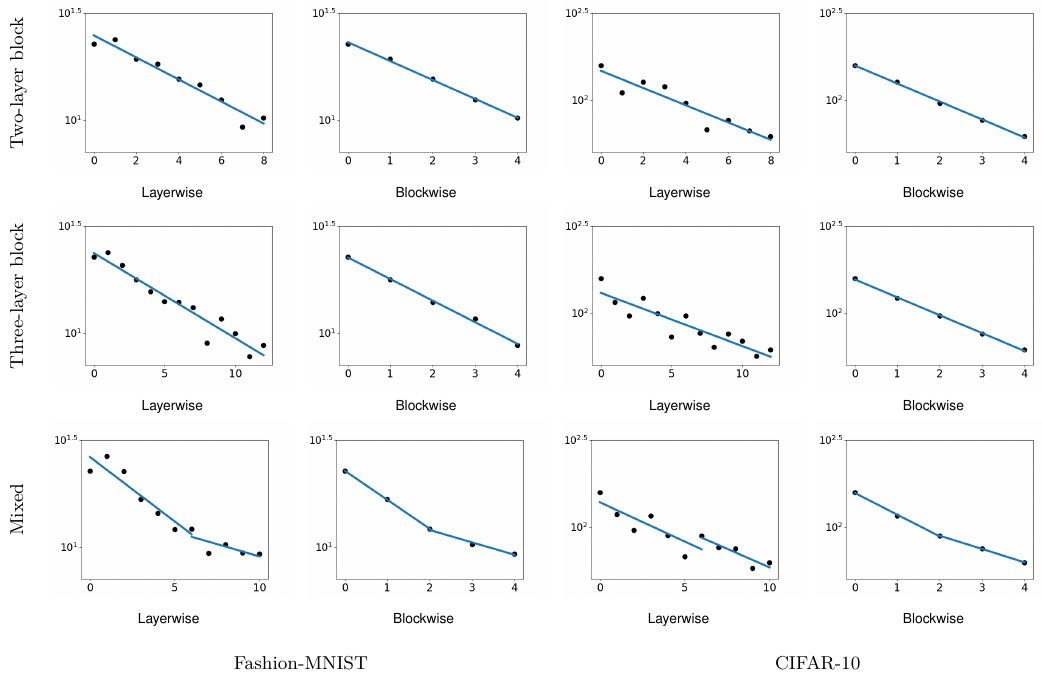}
    \caption{Illustration of the law of equi-separation in residual neural networks: Each block contains either two or three layers. In the ``Mixed'' configuration, the first two blocks consist of three layers each, while the last two blocks have two layers each. In each plot, a layer or a block is identified as a module when presenting the $x$ axis. The first two columns are evaluated on the Fashion-MNIST dataset, and the last two columns are evaluated on the CIFAR-10 dataset.} 
			\label{fig:resnet_arc}
\end{figure*}

\begin{figure}[!htp]
	\centering
	\includegraphics[scale=0.95]{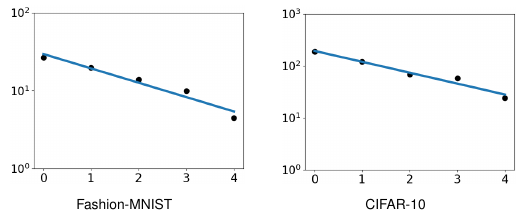}
    \caption{Illustration of the law of equi-separation in densely connected convolutional networks (DenseNet161) by identifying a block as a module. }
	\label{fig:DenseNet}
\end{figure}

This condition is precisely the case when the equi-separation law holds. Thus, the data separation ability of the classifier is not very sensitive to perturbations in network weights when the law emerges. Therefore, to improve the robustness, we need to train a neural network at least until the law of equi-separation comes into effect. In the literature, however, connections with robustness are often established through loss functions rather than the data separation perspective~\cite{bubeck2021universal}.

This proposition suggests that the law of equi-separation could be considered a demonstration of the neural networks' ``homogeneity'' across all layers. An essential component in the derivation of Proposition~\ref{prop} involves the assumption that the magnitude of the perturbation remains uniform across all layers, as seen from Eq.~\ref{eq:homo}. Even though it is improbable for the perturbation to remain constant in practice, it is intriguing to note that the emergence of this law is contingent on the use of batch normalization \citep{ioffe2015batch}, which enhances the homogeneity of the networks in a certain sense. Indeed, 
Fig.~S10 in SI Appendix shows that the law does not appear when batch normalization is not used.

\begin{figure}[!htp]
	\centering
	\includegraphics[scale=0.95]{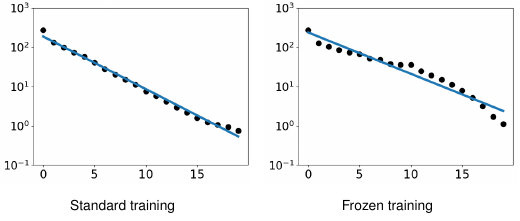}
    \caption{Illustration showing the law of equi-separation manifesting in the left network but not in the right network because of frozen training. Both networks have about the same training losses as well as training accuracies. However, the left network has a higher test accuracy (23.85\%) than that of the right network (19.67\%).}
		\label{fig:law-generalization_main}
\end{figure}

The law of equi-separation can also shed light on the out-of-sample performance of neural networks. Fig.~\ref{fig:law-generalization_main} considers two neural networks where one exhibits the law of equi-separation and the other does not. The results of our experiments showed that the former one has better test performance. Although this law is sometimes not observed in neural networks with good test performance, interestingly, one can fine-tune the parameters to reactivate the law with about the same or better test performance.

Moreover, the law of equi-separation can be potentially used to facilitate the training of transfer learning. In transfer learning, for example, bottom layers trained from an upstream task can form part of a network that will be trained on a downstream task. A challenge arising in practice is how to balance between the training times on the source and the target. In light of the law of equi-separation, a sensible balance is to find when the improvement in reducing separation fuzziness is roughly the same between the bottom and top layers.

\subsection{Interpretation} 
\begin{figure}[!htp]
	\centering
	\includegraphics[scale=1.0]{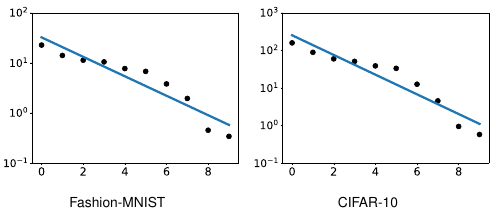}
    \caption{The law of equi-separation in ResNet18 is not as pronounced as it is in shallower residual neural networks.}
	\label{fig:ResNet18}
\end{figure}

The equi-separation law offers a new perspective for interpreting deep learning predictions, especially in high-stakes decision makings. An important ingredient in interpretation is to pinpoint the basic operational modules in neural networks and subsequently to reflect that ``all modules are created equal.'' For feedforward and convolutional neural networks, each layer is a module as it reduces the separation fuzziness by an equal multiplicative factor 
(for more details, see Fig.~S11 in SI Appendix).
Interestingly, even when the law no longer manifests in residual neural networks, Fig.~\ref{fig:resnet_arc} shows that it can be restored by identifying a block rather than a layer as a module. This figure also suggests that blocks containing more layers are more capable of reducing the separation fuzziness. Likewise, Fig.~\ref{fig:DenseNet} shows that the law of equi-separation holds in densely connected convolutional networks by identifying a block as a module. For comparison, when the perspective of data separation is not considered~\cite{JMLR:v23:20-069}, the correct module cannot be identified. For completeness, it should be noted that the law becomes less clear for deeper residual neural networks\footnote{See the architecture in \url{https://github.com/kuangliu/pytorch-cifar/tree/master/models}.}, as shown in Fig.~\ref{fig:ResNet18}.

Because each module makes equal but small contributions, a few neurons in a single layer are unlikely to explain how a deep learning model makes a particular prediction. It is therefore crucial to take all layers collectively for interpretation~\cite{su2021neurashed}. This view, however, challenges the widely used layer-wise approaches to deep learning interpretation~\cite{zeiler2014visualizing,tenney2019bert}. 

\section{Discussion}

Due to \cite{papyan2020prevalence} and follow-up work, it has been known that neural networks can exhibit precise mathematical structures at the last layer in the terminal phase of training. In this paper, we have extended this viewpoint from the surface of these black-box models to their interior by introducing an empirical law that quantitatively governs how well-trained real-world neural networks separate data throughout all layers. This law has been demonstrated to provide useful guidelines and insights into deep learning practice through network architecture design, training, and interpretation of predictions.

One avenue for future research is to explore the applicability of the law of equi-separation to a wider range of network architectures and applications. For example, it would be interesting to see if this law holds for neural ordinary differential equations and, if so, how it could be used to improve our understanding of the depth of these continuous-depth models. Additionally, it would be valuable to examine the law in regression settings and multi-modal tasks~\cite{liang2022mind}. A crucial question that remains to be addressed is understanding why the law is most accurate in feedforward neural networks and becomes somewhat fuzzier in certain convolutional neural networks. Next, given that the law is most evident in feedforward neural networks, an interesting direction is to explore new measures in lieu of the separation fuzziness defined in Eq.~\ref{eq:trace}. The hope is that this could potentially make the law clearer in other network architectures. In doing so, it would be beneficial to take into account network structures, such as convolution kernels in convolutional neural networks, during the development of new separation measures. Furthermore, while the law of equi-separation has been observed in a variety of vision tasks, it does not appear to hold for language models such as BERT~\cite{kenton2019bert} 
(see Fig.~S9 in SI Appendix).
This could be due to the fact that language models learn token-level features rather than sentence-level features at each layer. As such, it would be important to study whether the law of equi-separation could be utilized to derive better sentence-level features from a sequence of token-level features.

Perhaps the most pressing question is to reveal the underlying mechanism of this data-separation law. However, some commonly used assumptions in the literature of deep learning theory may not be applicable. For instance, the nonlinearity of activation functions is crucial to this law because the separation fuzziness is roughly unchanged by linear transformations. Additionally, analyses of neural networks using the neural tangent kernel~\citep{jacot2018neural} are not able to capture the effect of depth, which is essential for understanding the data separation process across all layers. This suggests that more innovative approaches may be necessary to fully understand the mechanisms behind this law.

Broadly speaking, a perspective informed by this law is to focus on the non-temporal dynamics of the data separation process throughout all layers when studying deep neural networks. This perspective probes into the interior of deep learning models by perceiving the black-box classifier as a sequence of transformations, each of which corresponds to a layer or a block in the case of residual neural networks. In contrast, existing analyses often consider the ``exterior'' of these models by focusing on the loss function or test performance, without considering the intermediate layers~\cite{jacot2018neural,he2019local,belkin2019reconciling}. Informed by the law of equi-separation, this perspective may provide formal insights into the inner workings of deep learning.

\subsection*{Data Availability} 
Our code is  publicly  available  at  GitHub (\url{https://github.com/HornHehhf/Equi-Separation}).

\showmatmethods{} 

\acknow{We are grateful to X.Y.~Han for very helpful comments and feedback on an early version of the manuscript. We also thank Hangyu Lin (\url{https://github.com/avalonstrel/NeuralCollapse/tree/mlp}), Cheng Shi (\url{https://github.com/DaDaCheng/Re-equi-sepa}), and Ben Zhou (\url{https://github.com/Slash0BZ/data-separation/tree/main/reproduction}) for reproducing our experimental results. This research was supported in part by NSF grants CCF-1934876 and CAREER DMS-1847415, and an Alfred Sloan Research Fellowship.}

\showacknow{} 


\bibliography{ref}

\clearpage

\section*{Supporting Information}
\label{sec:supp-inform}

\setcounter{figure}{0}
\renewcommand{\figurename}{Fig.}
\renewcommand{\thefigure}{S\arabic{figure}}



In this appendix, we first detail the general setup for all experiments, then briefly highlight some distinctive settings for figures in the main text, after that mention the argument for robustness, and finally show some additional results. More details can be found in our code.\footnote{Our code is  publicly  available  at  GitHub (\url{https://github.com/HornHehhf/Equi-Separation}).}

\subsection*{General Setup} 
\label{subsec:setup}

In this subsection, we detail the general setup that is used for the experiments throughout the paper.

\paragraph{Network architecture.} In our experiments, we mainly consider feedforward neural networks (FNNs). Unless otherwise stated, the corresponding hidden size is $100$. For simplicity, we use neural networks with $4$, $8$, $20$ layers as representatives for shallow, mid, and deep neural networks. By default, batch normalization \citep{ioffe2015batch} is inserted after fully connected layers and before the ReLU nonlinear activation functions.

\paragraph{Dataset.} Our experiments are mainly conducted on Fashion-MNIST. Unless otherwise stated, we resize the original images to $10 \times 10$ pixels. By default, we randomly sample a total of $1000$ training examples evenly distributed over the $10$ classes.

\paragraph{Optimization methodology.} In our experiments, three popular optimization methods are considered: SGD,  SGD with momentum, and Adam. The weight decay is set to $5 \times 10^{-4}$, and the momentum term is set to $0.9$ for SGD with momentum. The neural networks are trained for $600$ epochs with a batch size of $128$. The initial learning rate is annealed by a factor of $10$ at $1/3$ and $2/3$ of the $600$ epochs. As for SGD and SGD with momentum, we pick the  model  resulting in  the  best equi-separation law in the last epoch among the seven learning rates: $0.001$, $0.003$, $0.01$, $0.03$, $0.1$, $0.3$, and $1.0$. Similarity, we pick the model resulting in the best equi-separation law in the last epoch for Adam among the five learning rates: $3\times10^{-5}$, $1 \times 10^{-4}$, $3 \times 10^{-4}$, $1 \times 10^{-3}$, $3 \times 10^{-3}$. Unless otherwise stated, Adam is adopted in the experiments.

\subsection*{Detailed Experimental Settings} 
\label{subsec:experimental-details}

In this subsection, we show detailed experimental settings for the figures in the main text.

\paragraph{Optimization.} As shown in Fig.~\ref{fig:optimization}, for FNNs with different depths, three different optimization methods are considered: SGD, SGD with momentum, and Adam.

\paragraph{Visualization.} As shown in Fig.~\ref{fig:optimization}, we visualize the features of different layers in the 8-layer FNNs trained on Fashion-MNIST. In particular, principal component analysis (PCA) is used to visualize the features in a two-dimensional plane. 

\begin{figure}[!htp]
\centering
        \captionsetup[subfigure]{labelformat=empty}
		\subfloat[Epoch]{
			\centering
			\includegraphics[scale=0.45]{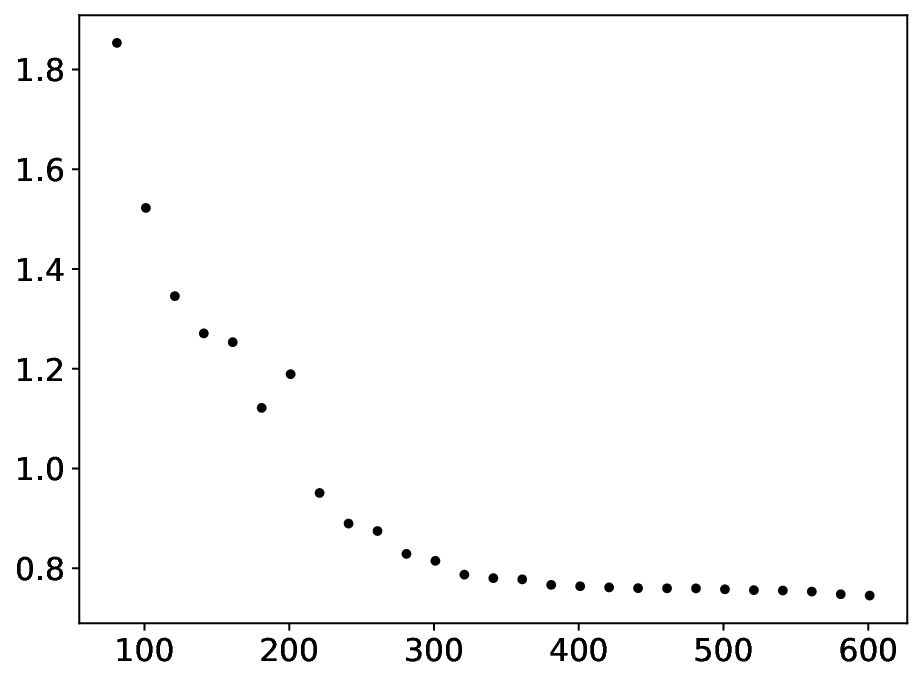}}
    \caption{Separation fuzziness of last-layer features throughout the training process. The $x$ axis represents the epoch number, while the  $y$ axis indicates the separation fuzziness ($D$ in Eq.~\ref{eq:trace}). The $x$ axis commences from approximately 100 epochs due to the exceedingly high initial separation fuzziness, which would otherwise mask discernible trends.}
		\label{fig:collapse-scores}
\end{figure}

\paragraph{Training epoch.} As shown in Fig.~\ref{fig:training-epoch}, we show the separability of features among layers in different training epochs. In this part, we simply consider the $20$-layer FNNs trained on Fashion-MNIST. The equi-separation law doesn't emerge at the beginning, and gradually emerges during training. After that the decay ratio is decreased along the training process until the network converges. Furthermore, we conducted an analysis of the last-layer features in the aforementioned experiments. As illustrated in Fig.~\ref{fig:collapse-scores}, it is evident that the separation fuzziness of the last-layer features does not reach convergence until after $400$ epochs. Conversely, the equi-separation law manifests at epoch $100$, as demonstrated in Fig.~\ref{fig:training-epoch}, with a Pearson correlation coefficient of $-0.996$. This observation indicates that the equi-separation law emerges prior to the occurrence of neural collapse.
\\

\noindent As shown in Fig.~\ref{fig:elaborate}, we further consider the impact of the following factors on the equi-separation law: dataset, learning rate, and class distribution.  

\paragraph{Dataset.} As shown in Fig.~\ref{fig:elaborate}, we experiment with CIFAR-10. We use the second channel of the images and resize the original images to $10 \times 10$. Similar to Fashion-MNIST, we randomly sample a total of $1000$ training examples evenly distributed over the $10$ classes. 

\paragraph{Learning rate.} As shown in Fig.~\ref{fig:elaborate}, we compare the $9$-layer FNNs trained using SGD with different leaning rates on Fashion-MNIST.

\paragraph{Imbalanced data.} As shown in Fig.~\ref{fig:elaborate}, we experiment with imbalanced data. Specifically, we consider $5$ majority classes with $500$ examples in each class, and $5$ minority classes with $100$ examples in each class. The law of equi-separation also emerges in neural networks trained with imbalanced Fashion-MNIST data, though its half-life is significantly larger than that in the balanced case. This might be caused by the collapse of minority classifiers as shown in \cite{fang2021exploring}.

\paragraph{Convolutional neural networks (CNNs).} As shown in Fig.~\ref{fig:CNNs}, we experiment with two canonical CNNs, AlexNet and VGG \citep{simonyan2014very}. We use the PyTorch implementation of both models.\footnote{More details are in \url{https://github.com/pytorch/vision/tree/main/torchvision/models}.}  We experiment with both Fashion-MNIST and CIFAR-10, and the original images are resized to $32 \times 32$ pixels. Given that AlexNet and VGG are designed for ImageNet images with $224 \times 224$ pixels, we make some small modifications to handle images with $32 \times 32$ pixels. As for AlexNet, we change the original convolution filters to $5 \times 5$ convolution filters with padding $2$. The color channel number is multiplied by $32$ at the beginning. As for VGG, we change the original convolution filters to $5 \times 5$ convolution filters with padding $2$, and the last average pooling layer before fully connected layers is replaced by an average pooling layer over a $1 \times 1$ pixel window. The color channel number is multiplied by $16$ at the beginning. 

\begin{figure*}[!htp]
	\centering
	\captionsetup[subfigure]{labelformat=empty}
        \centering
       \rotatebox[y=0.8cm]{90}{Fashion-MNIST}\quad
        \subfloat[VGG11]{
			\centering
	   \includegraphics[scale=0.24]{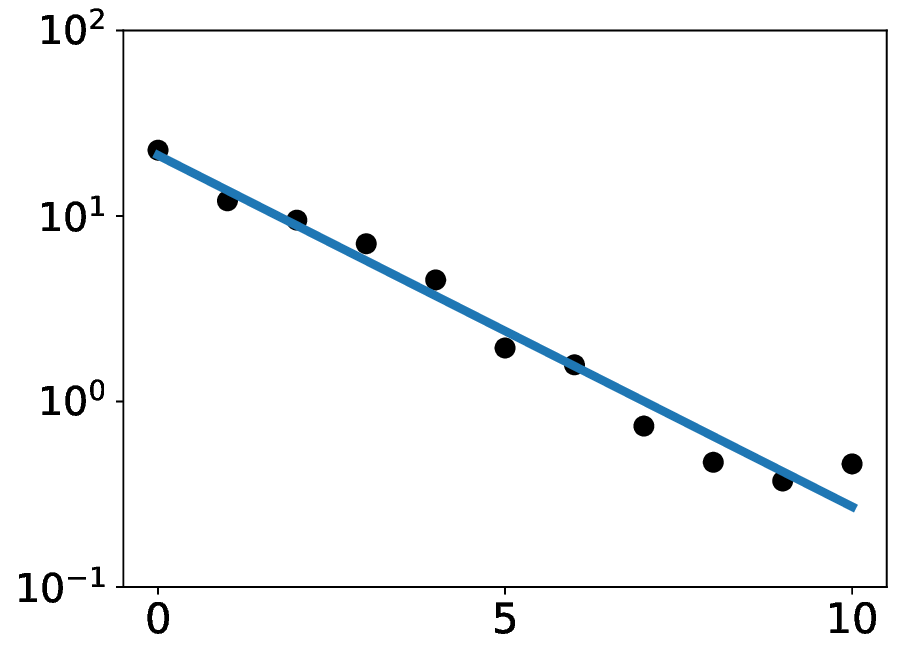}
			\label{fig:fashion-mnist-vgg11}}\hfill
        	\subfloat[VGG13]{
			\centering
		\includegraphics[scale=0.24]{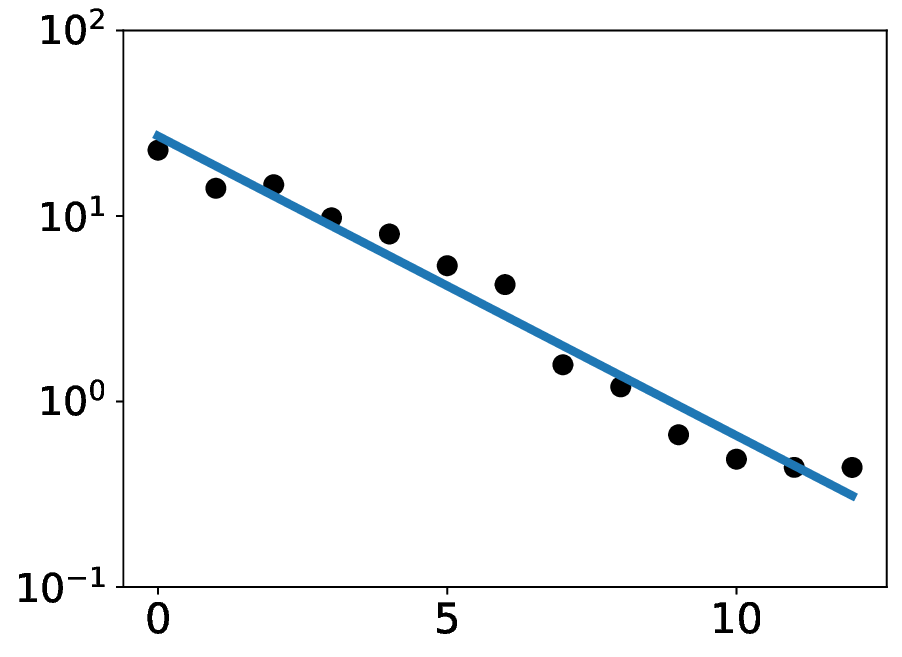}
			\label{fig:fashion-mnist-vgg13}}\hfill
     	\subfloat[VGG16]{
			\centering
		\includegraphics[scale=0.24]{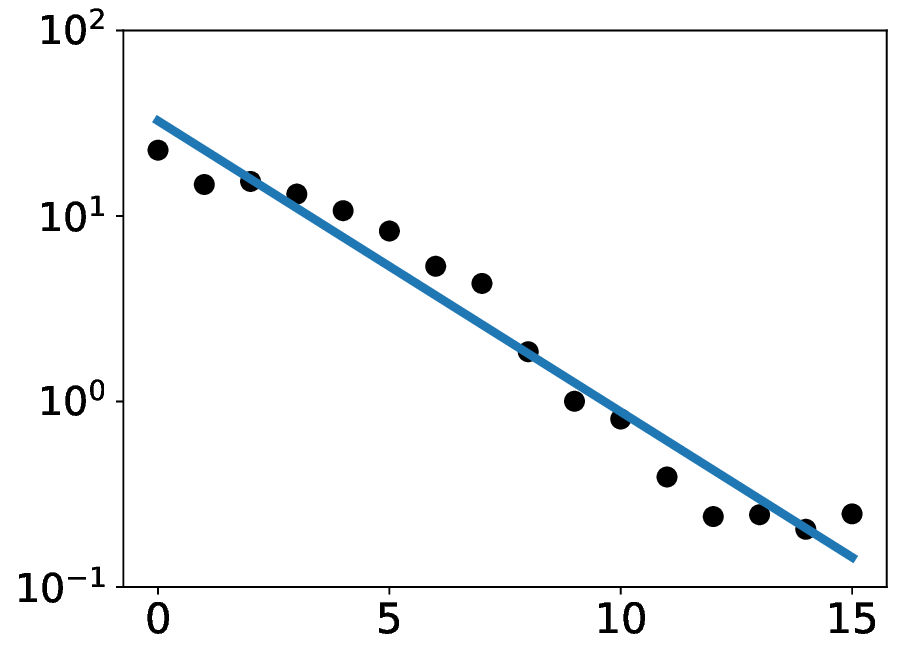}
			\label{fig:fashion-mnist-vgg16}}\hfill
         \subfloat[VGG19]{
			\centering
		\includegraphics[scale=0.24]{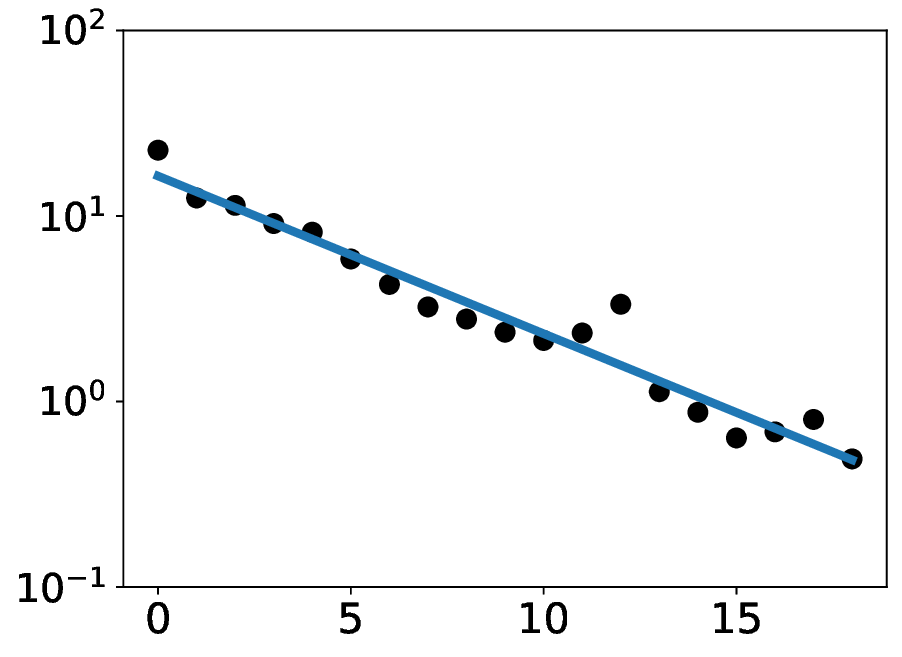}
			\label{fig:fashion-mnist-vgg19}}

    \rotatebox[y=1.0cm]{90}{CIFAR-10}\quad
		\subfloat[VGG11]{
			\centering
		\includegraphics[scale=0.24]{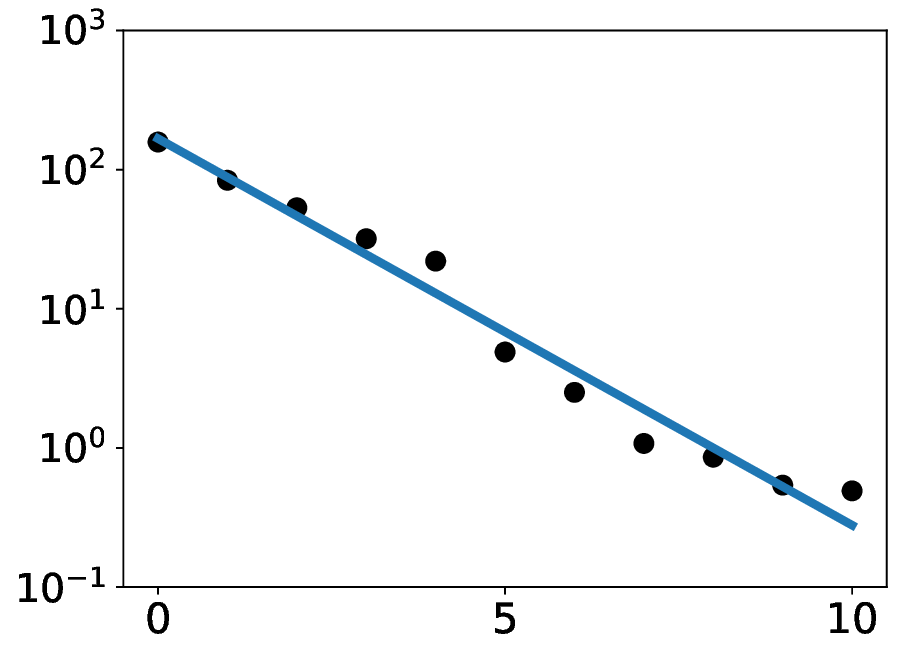}
			\label{fig:cifar10-vgg11}}\hfill
        	\subfloat[VGG13]{
			\centering
	 \includegraphics[scale=0.24]{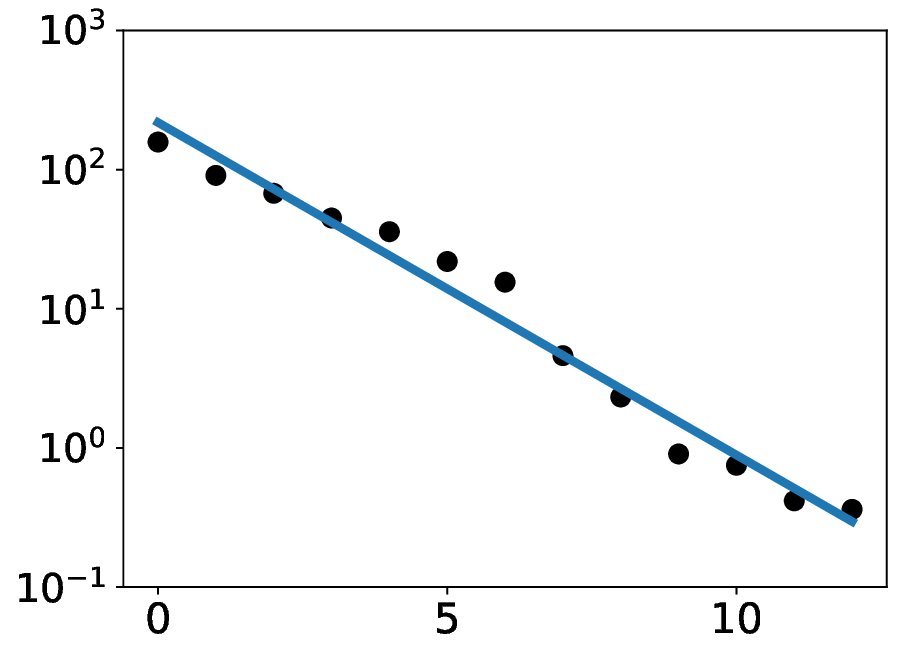}
			\label{fig:cifar10-vgg13}}\hfill
     	\subfloat[VGG16]{
			\centering
		\includegraphics[scale=0.24]{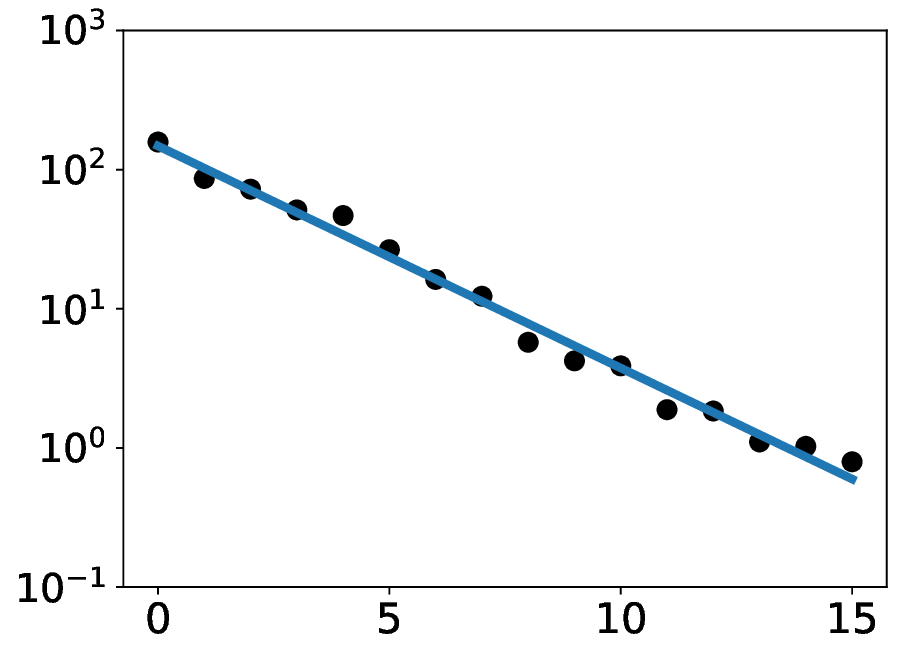}
			\label{fig:cifar10-vgg16}}\hfill
          \subfloat[VGG19]{
			\centering
	  \includegraphics[scale=0.24]{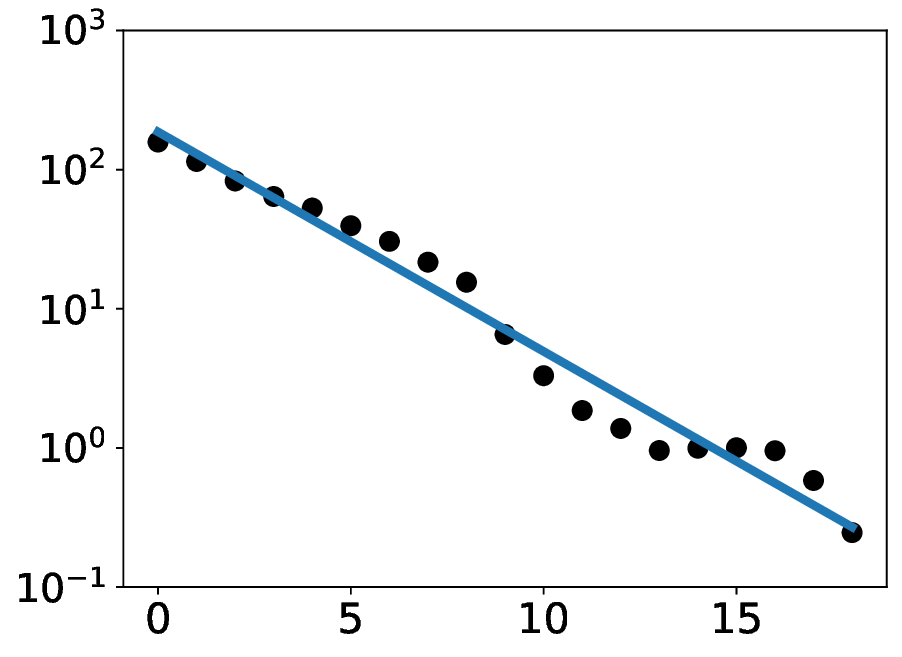}
			\label{fig:cifar10-vgg19}}
			 
    \caption{Emergence of a fuzzy equi-separation law in VGG networks with varied depths on Fashion-MNIST and CIFAR-10. }
	\label{fig:VGG}
\end{figure*}

Moreover, we examined VGG employing four distinct configurations\footnote{More details can be found at: \url{https://github.com/kuangliu/pytorch-cifar/tree/master/models}.} on the Fashion-MNIST and CIFAR-10 datasets. As demonstrated in Fig.~\ref{fig:VGG}, an approximate equi-separation law is also observed across different VGG configurations. It is important to note that the equi-separation law can be further refined by modifying convolution filter size and input channel number, as illustrated in Fig.~\ref{fig:CNNs}.

\begin{figure*}[!htp]
	\centering
	\captionsetup[subfigure]{labelformat=empty}
        \centering
       \rotatebox[y=0.8cm]{90}{Fashion-MNIST}\quad
        \subfloat[16 Channels]{
			\centering
	   \includegraphics[scale=0.24]{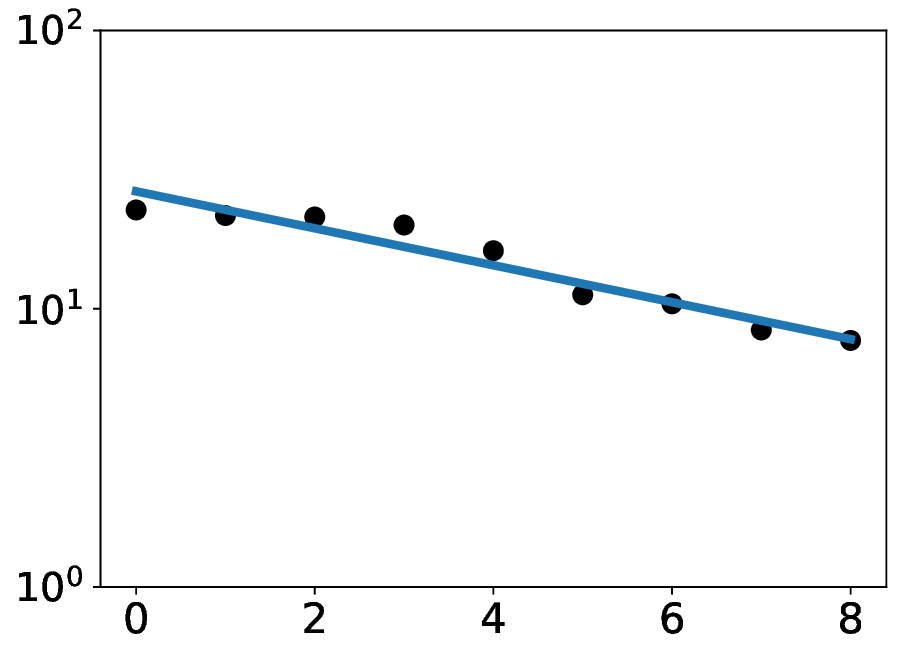}
			\label{fig:fashion-mnist-CNNL8C16}}\hfill
        	\subfloat[32 Channels]{
			\centering
		\includegraphics[scale=0.24]{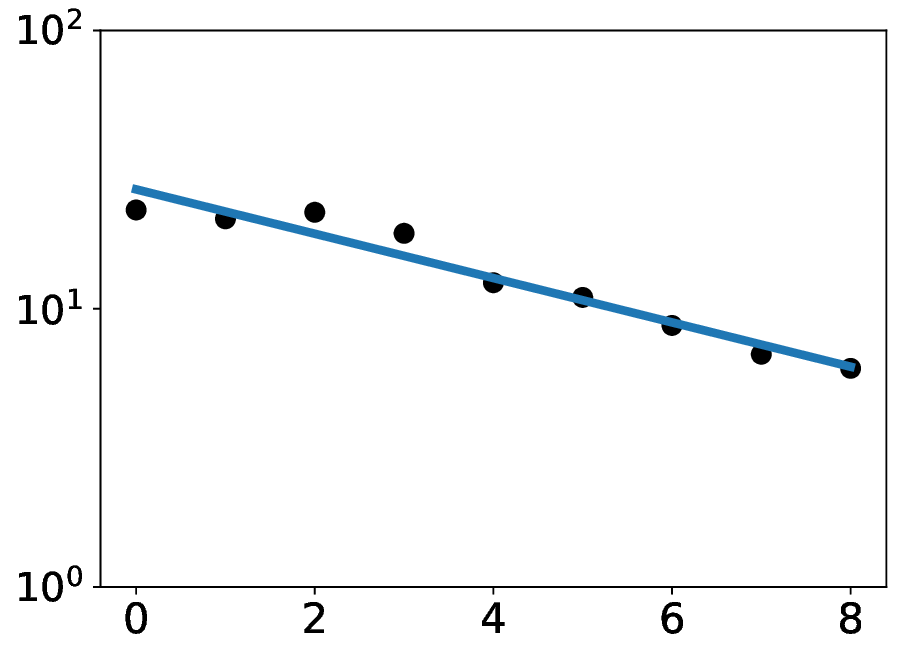}
			\label{fig:fashion-mnist-CNNL8C32}}\hfill
     	\subfloat[64 Channels]{
			\centering
		\includegraphics[scale=0.24]{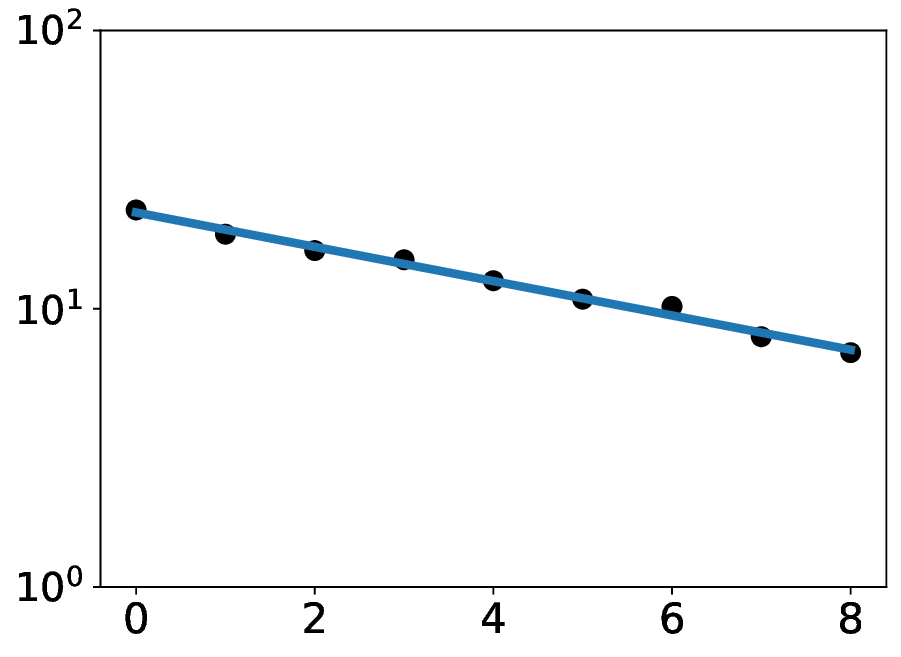}
			\label{fig:fashion-mnist-CNNL8C64}}\hfill
         \subfloat[128 Channels]{
			\centering
		\includegraphics[scale=0.24]{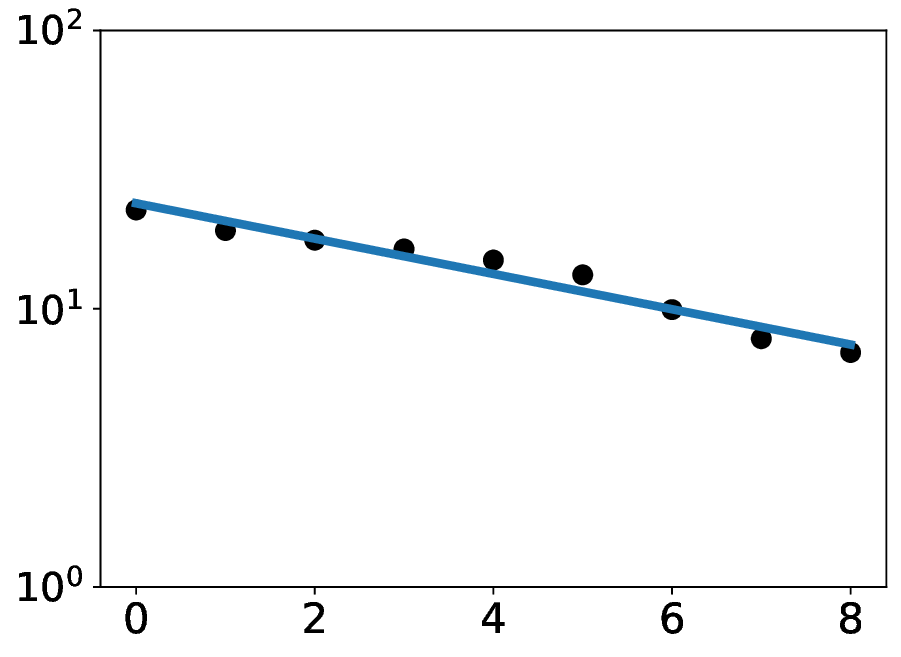}
			\label{fig:fashion-mnist-CNNL8C128}}

    \rotatebox[y=1.0cm]{90}{CIFAR-10}\quad
		\subfloat[16 Channels]{
			\centering
		\includegraphics[scale=0.24]{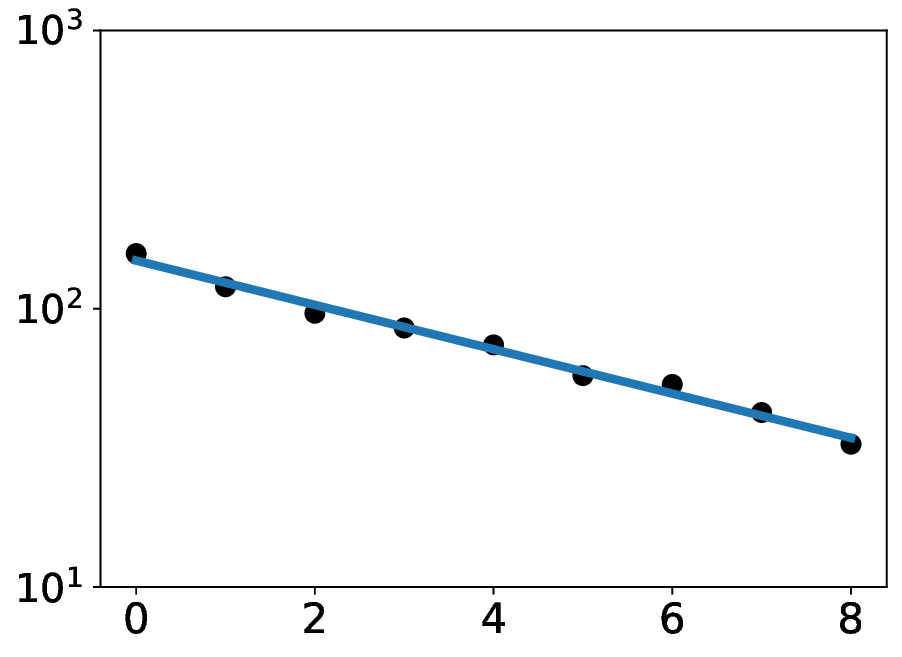}
			\label{fig:cifar10-CNNL8C16}}\hfill
        	\subfloat[32 Channels]{
			\centering
	 \includegraphics[scale=0.24]{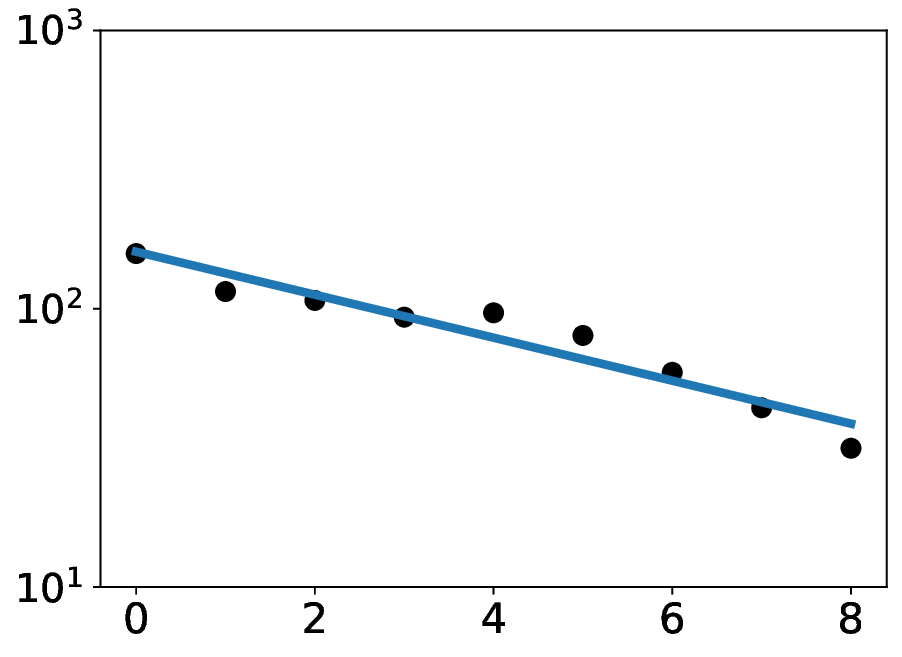}
			\label{fig:cifar10-CNNL8C32}}\hfill
     	\subfloat[64 Channels]{
			\centering
		\includegraphics[scale=0.24]{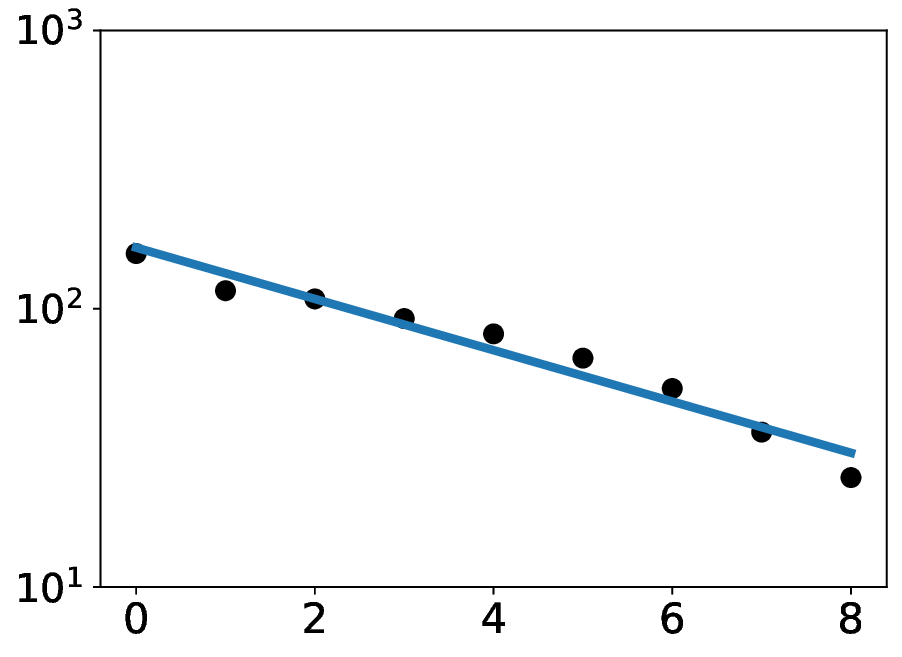}
			\label{fig:cifar10-CNNL8C64}}\hfill
          \subfloat[128 Channels]{
			\centering
	  \includegraphics[scale=0.24]{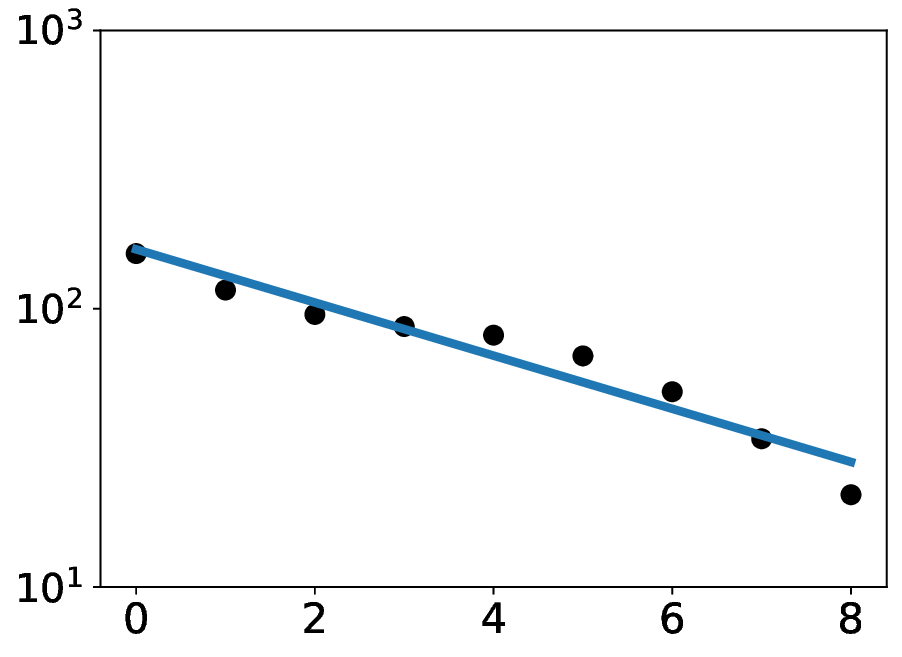}
			\label{fig:cifar10-CNNL8C128}}
			 
    \caption{Equi-separation law emerges in pure CNNs with appropriate numbers of channels on Fashion-MNIST (64 channels) and CIFAR-10 (16 channels). }
	\label{fig:pureCNNs}
\end{figure*}

To gain further insights into CNNs, we explored a variant of VGG11 (referred to as pure CNNs) by retaining all convolutional layers, the final fully connected layer, and removing all pooling layers along with the initial two fully connected layers. Additionally, we ensured an equal number of channels in each layer for our analysis. In this simplified network, convolutional layers serve to extract features, which are subsequently input into the final fully connected layer for classification. As presented in Fig.~\ref{fig:pureCNNs}, the equi-separation law is present in pure CNNs for various datasets with an appropriate number of channels; however, the separation fuzziness of the last-layer features is considerably higher than that of FNNs and canonical CNNs (e.g., AlexNet and VGG). We further investigated the influence of other factors in CNNs, such as depth, kernel size, pooling layer, and image size. Our findings indicate that the equi-separation law is indeed present in CNNs but necessitates a delicate balance among these factors. This suggests that an alternative measure, rather than separation fuzziness ($D$ in Eq.~\ref{eq:trace}), may be required to describe the information flow in CNNs, given the unique structure of features following convolutional layers.

\paragraph{Depth.} As shown in Fig.~\ref{fig:layer-number-dataset}, besides Fashion-MNIST and CIFAR-10, we also consider MNIST to illustrate the optimal depth for different datasets. Similar to Fashion-MNIST, we resize the original MNIST images to $10 \times 10$ pixels, and randomly sample a total of $1000$ training examples evenly distributed over the $10$ classes. 
\\

\noindent As shown in Fig.~\ref{fig:architectures}, we consider FNNs with different widths and shapes.

\paragraph{Width.} As shown in Fig.~\ref{fig:architectures}, we consider FNNs with three different widths. All layers have the same width in each setting. For simplicity, we consider the $8$-layer FNNs trained on Fashion-MNIST.  

\paragraph{Shape.} As shown in Fig.~\ref{fig:architectures}, we consider three different shapes of FNNs: Narrow-Wide, Wide-Narrow, and Mixed. As for Narrow-Wide, the FNNs have narrow hidden layers at the beginning, and then have wide hidden layers in the later layers.  Similarity, we use Wide-Narrow to denote FNNs that start with wide hidden layers followed by narrow hidden layers. In the case of Mixed, the FNNs have more complicated patterns of the widths of hidden layers.  For simplicity, we only consider $8$-layer FNNs trained on Fashion-MNIST. The corresponding widths of hidden layers for Narrow-Wide, Wide-Narrow, and Mixed we considered in our experiments are as follows: (100, 100,  100, 100, 1000, 1000, 1000), (1000, 1000, 1000, 1000, 100, 100, 100), and (100, 500, 500, 2500, 2500, 500, 500). 



\paragraph{Residual neural networks (ResNets).} As shown in Fig.~\ref{fig:resnet_arc}, we consider three different types of ResNets: ResNets with $2$-layer building blocks, ResNets with $3$-layer bottleneck blocks, and ResNets with mixed blocks. For $3$-layer blocks, the expansion rate is set to $1$ instead of $4$, and the original $1 \times 1$ and $3 \times 3$ convolution filters are replaced by $5 \times 5$ convolution filters. For all types of ResNets, the channel number is set to $8$, and the color channel number is also multiplied by $8$ at the beginning. In this part, the Fashion-MNIST and CIFAR-10 images are resized to $32 \times 32$ pixels instead of $10 \times 10$ pixels. 

\paragraph{Dense Convolutional Network (DenseNet).} As shown in Fig.~\ref{fig:DenseNet}, we investigate the law when applied to DenseNet \citep{huang2017densely}, namely DenseNet161.\footnote{Further details can be found at \url{https://github.com/kuangliu/pytorch-cifar/tree/master/models}.} In this part, the images of Fashion-MNIST and CIFAR-10 are resized to $32\times32$ pixels, i.e., $(3, 32, 32)$ for CIFAR-10 and $(32, 32)$ for Fashion-MNIST. 

\paragraph{Out-of-sample performance.} As shown in Fig.~\ref{fig:law-generalization_main}, given $20$-layer FNNs, instead of the standard training, we also consider the following two-stage frozen training procedure: 1) freeze the last $10$ layers and train the first $10$ layers; 2) freeze the first $10$ layers and train the last $10$ layers. When we conduct the frozen and standard training on CIFAR-10, we find that the our-of-sample performance of standard training (test accuracy: $23.85\%$) is better than that of frozen training (test accuracy: $19.67\%$), even though the two training procedures have similar training losses (standard: $0.0021$; frozen: $0.0019$) and training accuracies (standard: $100\%$; frozen: $100\%$). This is consistent with the equi-separation law: the neural networks trained using standard training have much clearer equi-separation law compared to the neural networks trained using frozen training. Specifically, the corresponding Pearson correlation coefficients for standard and frozen training are $-0.997$ and $-0.971$, respectively.

\subsection*{Additional Results}
\label{subsec:additional-results}

In this subsection, we provide some additional results for the equi-separation law.

\begin{figure*}[!htp]
	\centering
	\captionsetup[subfigure]{labelformat=empty}
        \centering
       \rotatebox[y=1.2cm]{90}{Fashion-MNIST}\quad
        \subfloat[Depth=4]{
			\centering
			\includegraphics[scale=0.33]{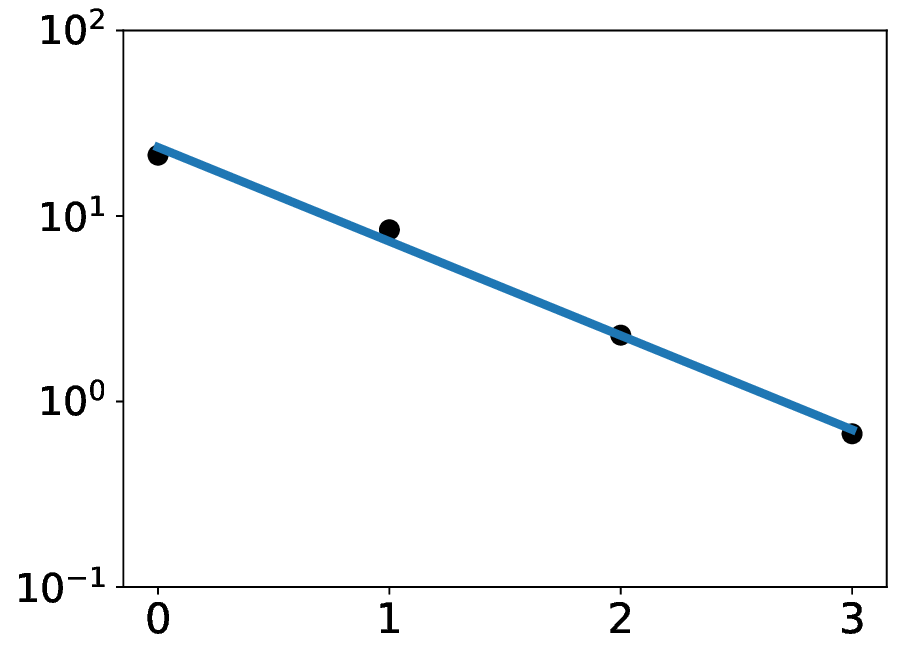}
			\label{fig:fashion-mnist-4-original}}\hfill
        	\subfloat[Depth=8]{
			\centering
			\includegraphics[scale=0.33]{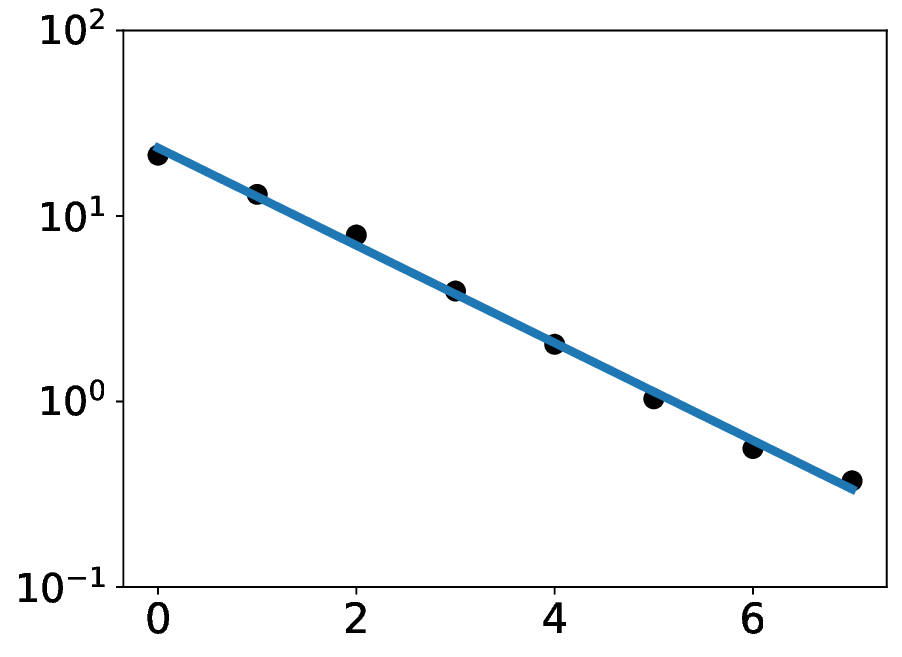}
			\label{fig:fashion-mnist-8-original}}\hfill
     	\subfloat[Depth=20]{
			\centering
			\includegraphics[scale=0.33]{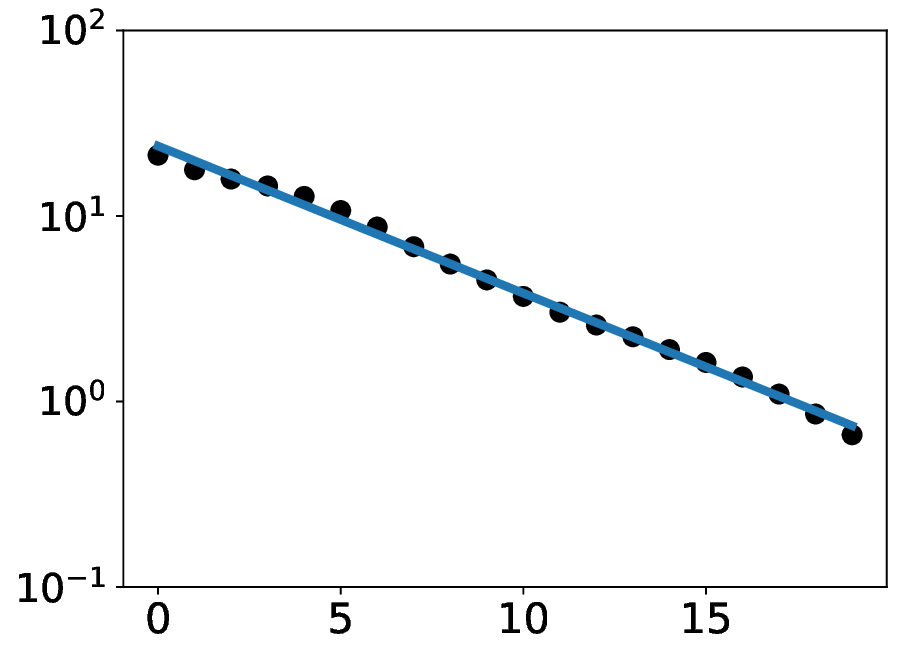}
			\label{fig:fashion-mnist-20-original}}

    \rotatebox[y=1.4cm]{90}{CIFAR-10}\quad
		\subfloat[Depth=4]{
			\centering
			\includegraphics[scale=0.33]{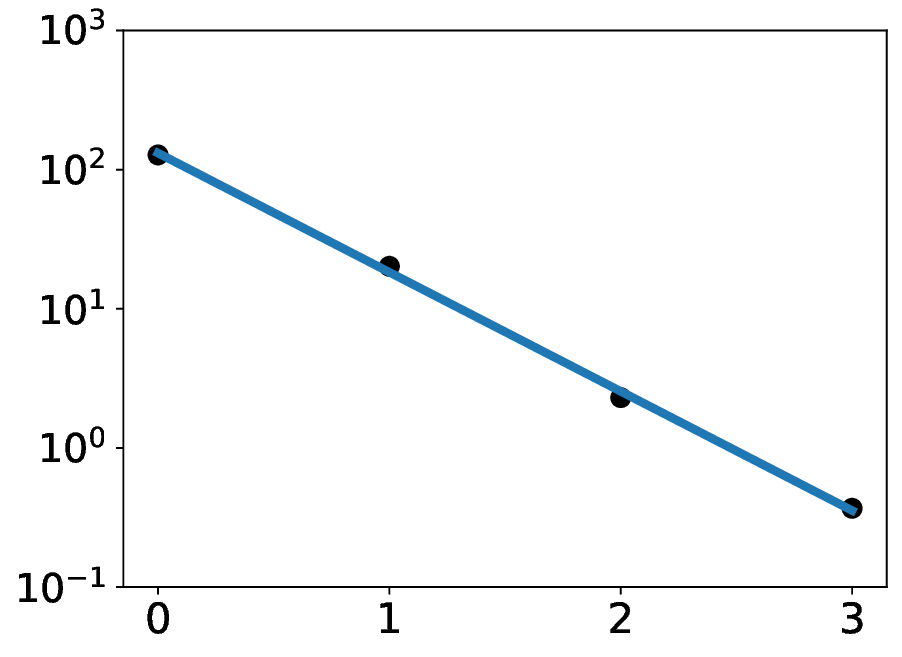}
			\label{fig:cifar10-4-original}}\hfill
        	\subfloat[Depth=8]{
			\centering
			\includegraphics[scale=0.33]{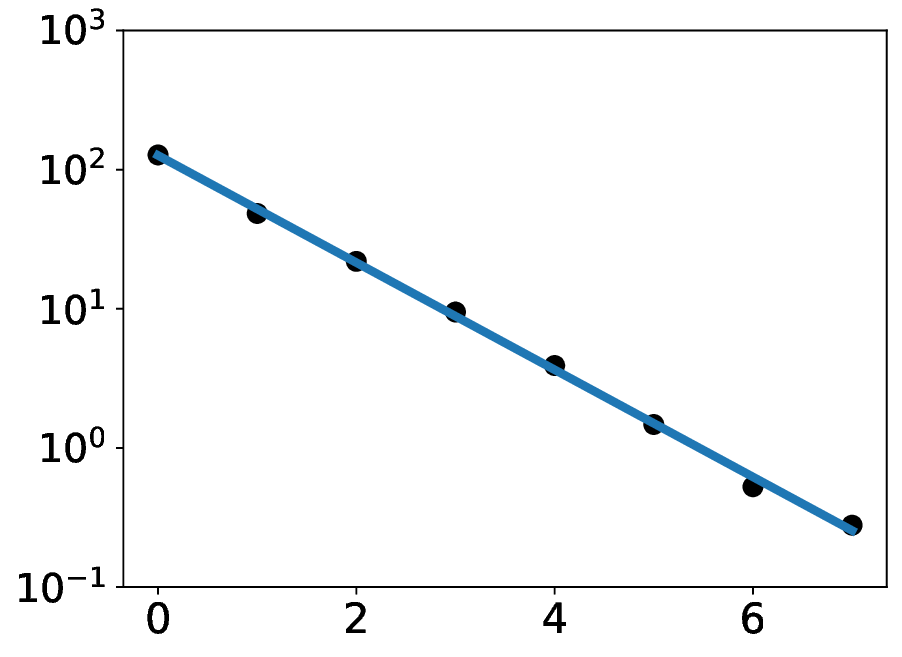}
			\label{fig:cifar10-8-original}}\hfill
     	\subfloat[Depth=20]{
			\centering
			\includegraphics[scale=0.33]{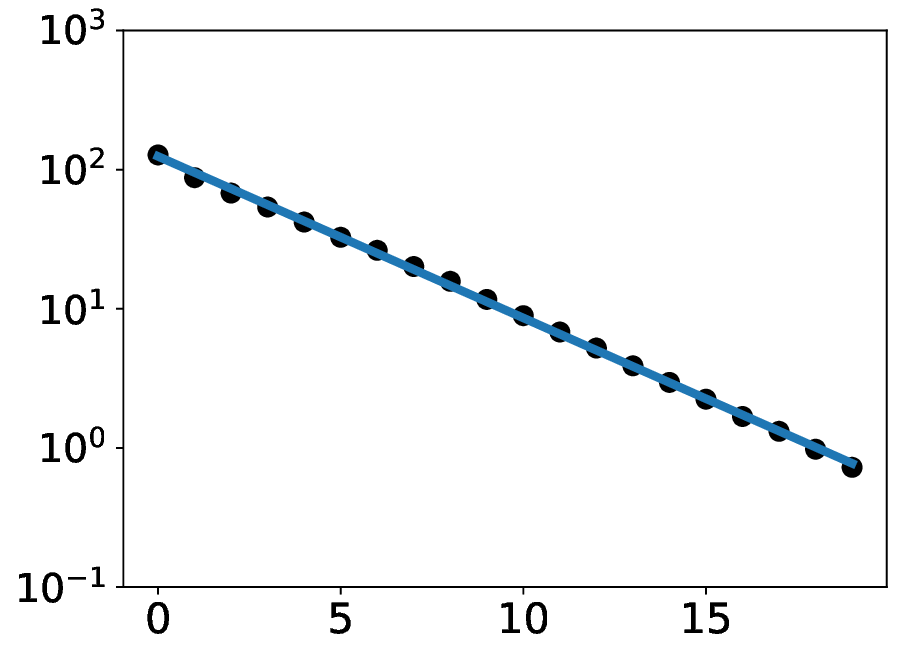}
			\label{fig:cifar10-20-original}}
			 
    \caption{Equi-separation law emerges in FNNs across varied depths on $32 \times 32$ pixel images of Fashion-MNIST and CIFAR-10. }
	\label{fig:original}
\end{figure*}

\paragraph{Image size.} As depicted in Fig.~\ref{fig:original}, we examine FNNs on images with a resolution of $32 \times 32$ pixels, specifically $(3, 32, 32)$ for CIFAR-10 and $(32, 32)$ for Fashion-MNIST. It is important to note that Fashion-MNIST, despite its original size of $(28, 28)$, is frequently resized to $(32, 32)$ in the same manner as MNIST. In this part, we set the width as $1000$ instead of $100$ for FNNs, given that the images with $32 \times 32$ pixel resolution possess a higher dimension in comparison to those with $10 \times 10$ pixel resolution.

\begin{figure*}[!htp]
         \captionsetup[subfigure]{labelformat=empty}
		\centering
		\subfloat[Sample size=1000]{
			\centering
			\includegraphics[scale=0.33]{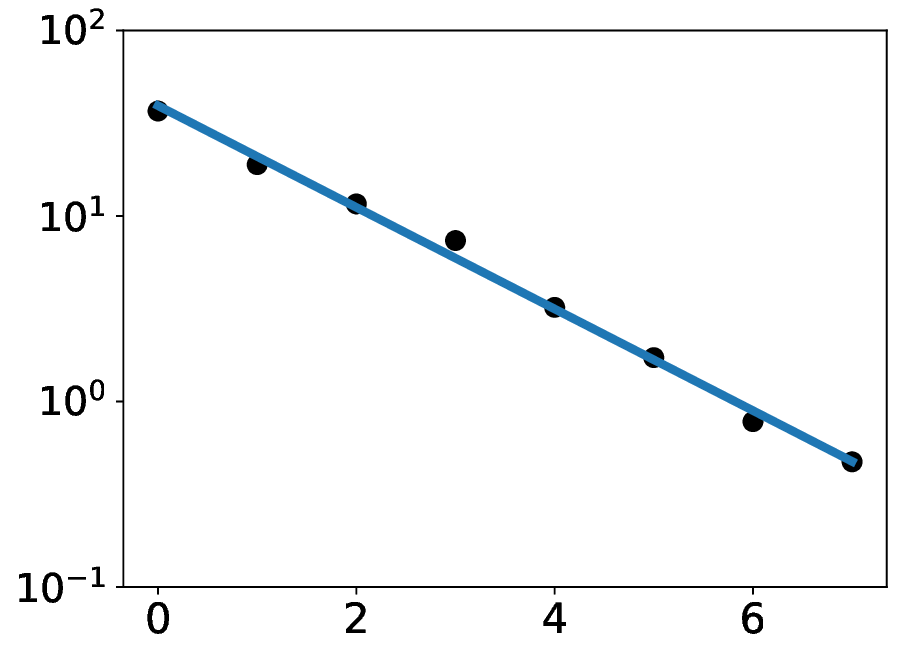}
			\label{fig:sample-100}}\hfill
        	\subfloat[Sample size=2000]{
			\centering
			\includegraphics[scale=0.33]{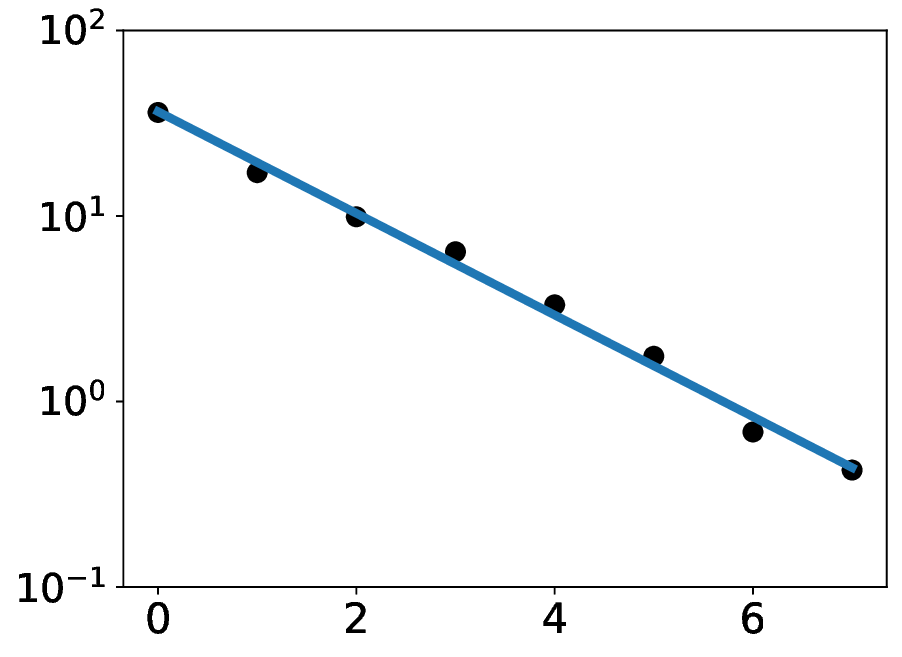}
			\label{fig:sample-200}}\hfill
     	\subfloat[Sample size=5000]{
			\centering
			\includegraphics[scale=0.33]{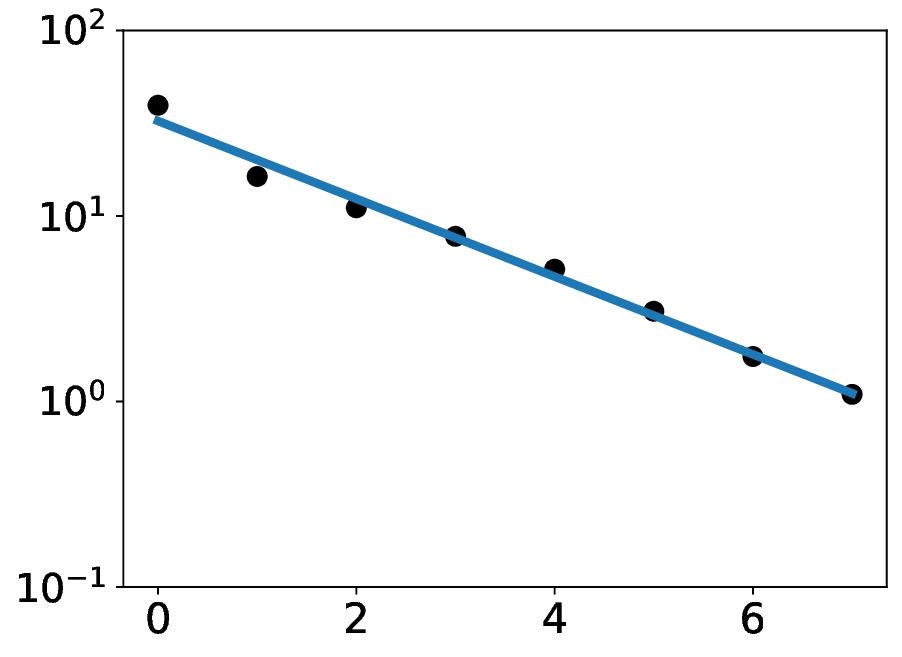}
			\label{fig:sample-500}}
			
		\caption{The impact of the sample size on the law of equi-separation.
		}
		\label{fig:sample-size}
\end{figure*}

\paragraph{Sample size.}  As shown in Fig.~\ref{fig:sample-size}, we experiment with different number of randomly sampled training examples. For each setting, we have the same number of examples for each of the $10$ classes. For simplicity, we only consider the $8$-layer FNNs trained on Fashion-MNIST. The law of equi-separation emerges in neural networks trained with different number of training examples, though the half-life can be larger for larger datasets due to the increased optimization difficulty.

\begin{figure*}[!htp]
	\centering
	\captionsetup[subfigure]{labelformat=empty}
        \subfloat[Class 1]{
			\centering
			\includegraphics[scale=0.19]{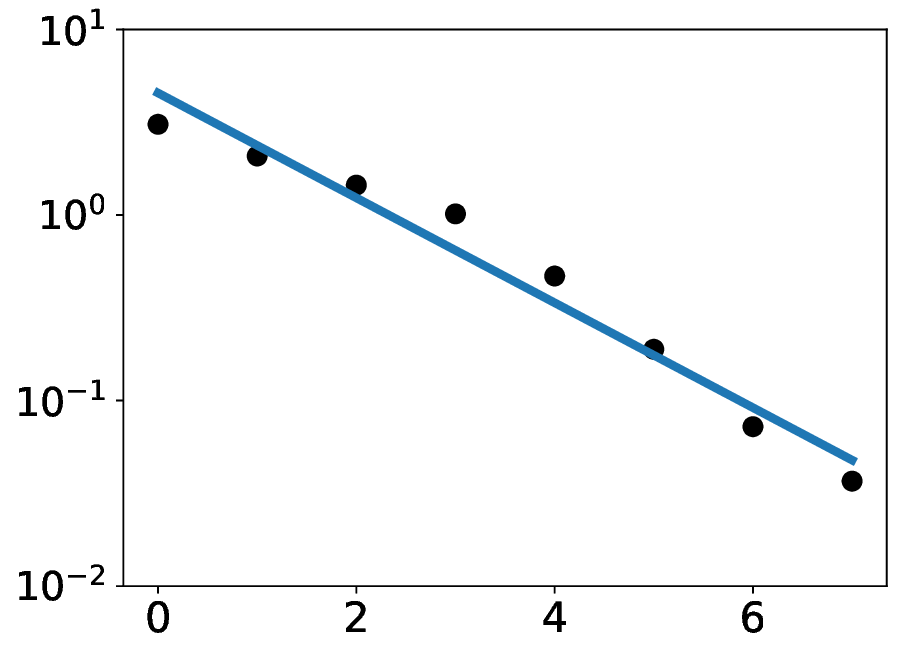}
			\label{fig:class-1}}\hfill
        \subfloat[Class 2]{
			\centering
			\includegraphics[scale=0.19]{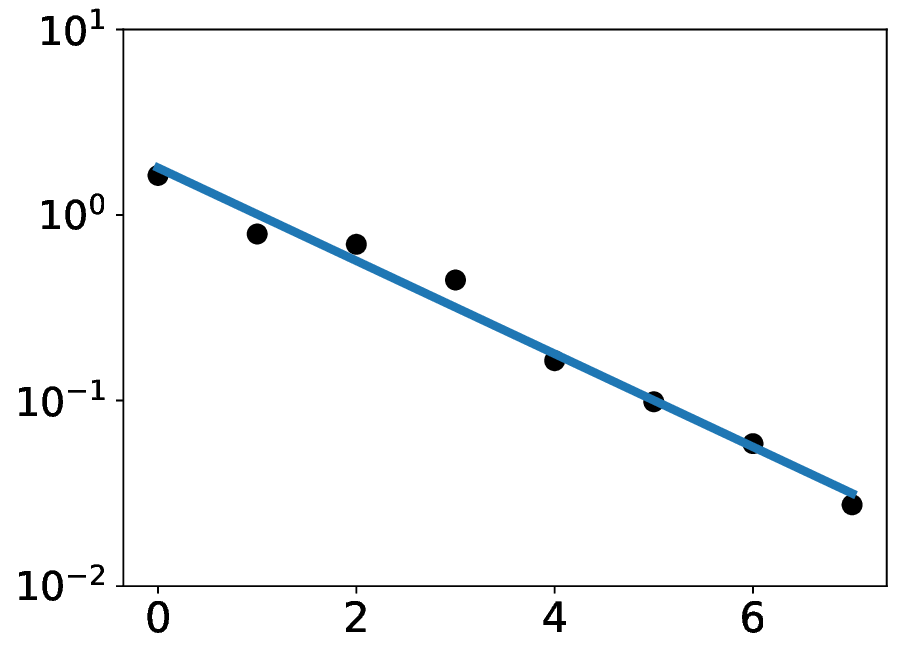}
			\label{fig:class-2}}\hfill
     	\subfloat[Class 3]{
			\centering
			\includegraphics[scale=0.19]{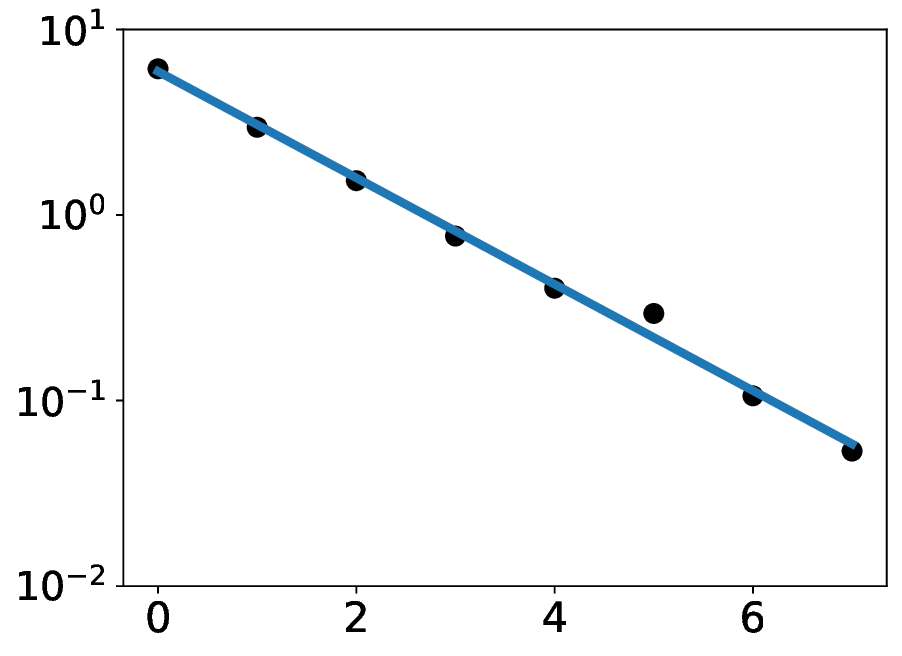}
			\label{fig:class-3}}\hfill
		\subfloat[Class 4]{
			\centering
			\includegraphics[scale=0.19]{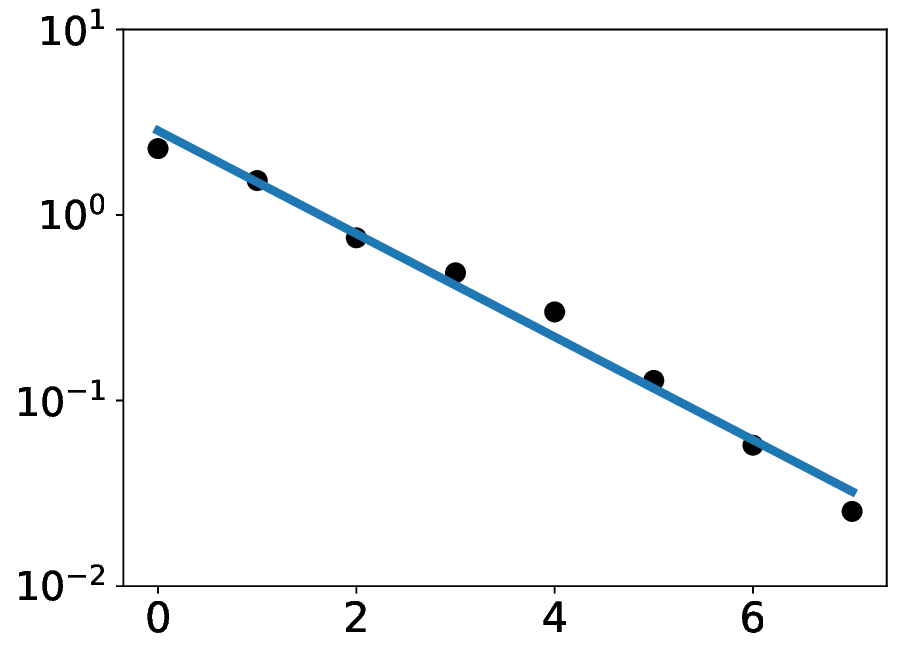}
			\label{fig:class-4}}\hfill
        \subfloat[Class 5]{
			\centering
			\includegraphics[scale=0.19]{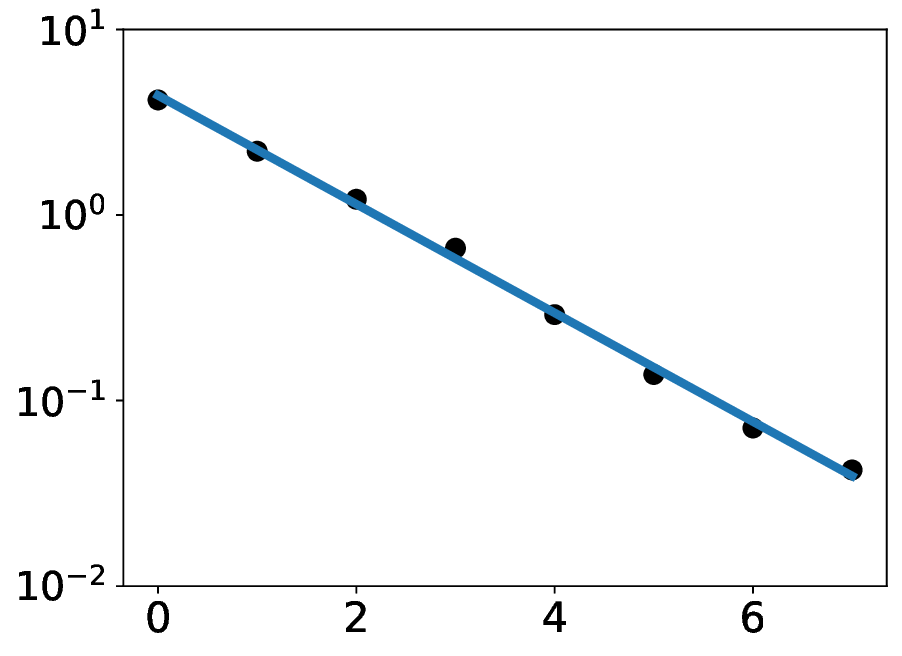}
			\label{fig:class-5}}
			
     	\subfloat[Class 6]{
			\centering
			\includegraphics[scale=0.19]{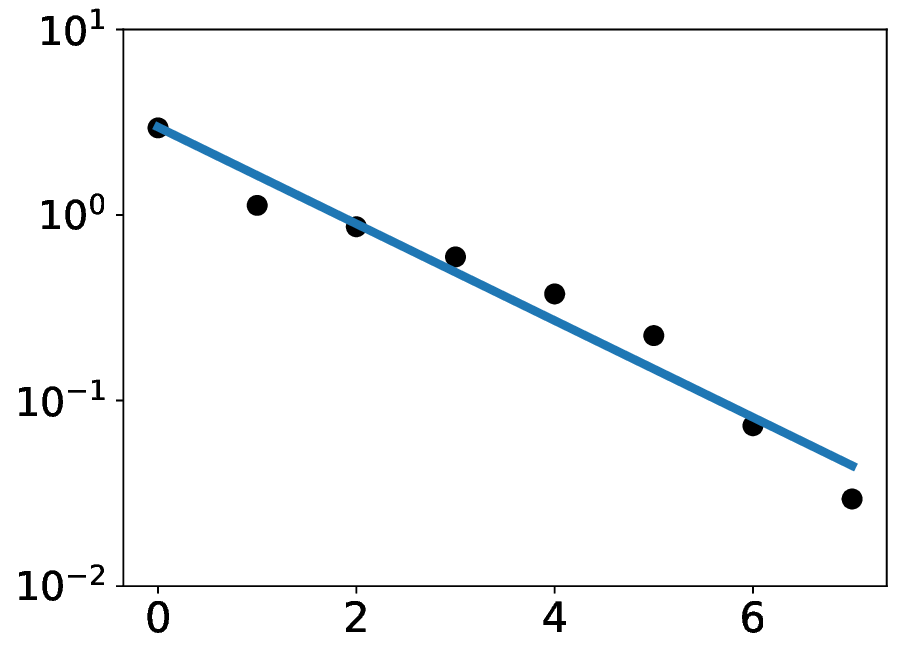}
			\label{fig:class-6}}\hfill
		\subfloat[Class 7]{
			\centering
			\includegraphics[scale=0.19]{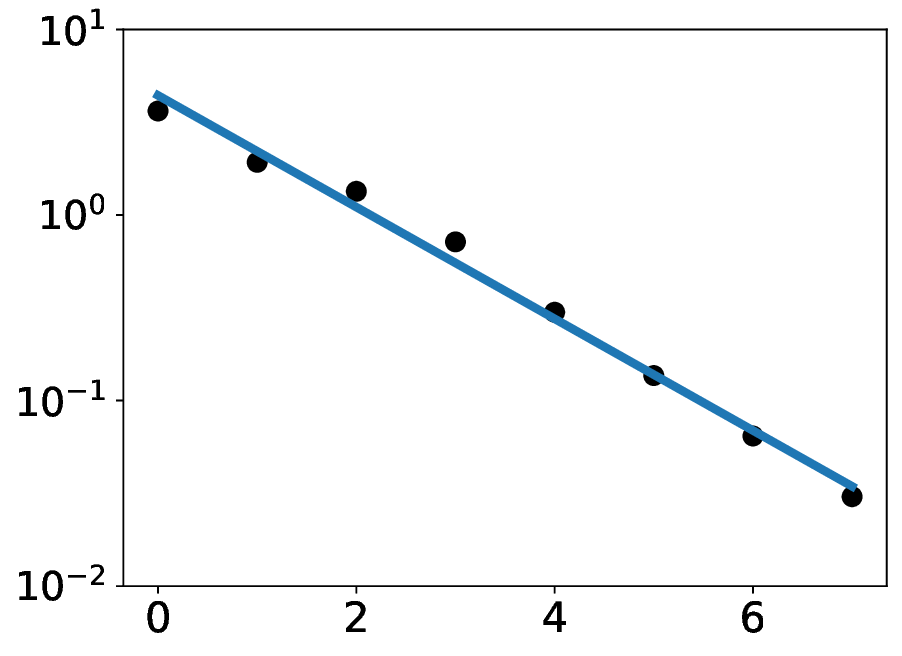}
			\label{fig:class-7}}\hfill
        \subfloat[Class 8]{
			\centering
			\includegraphics[scale=0.19]{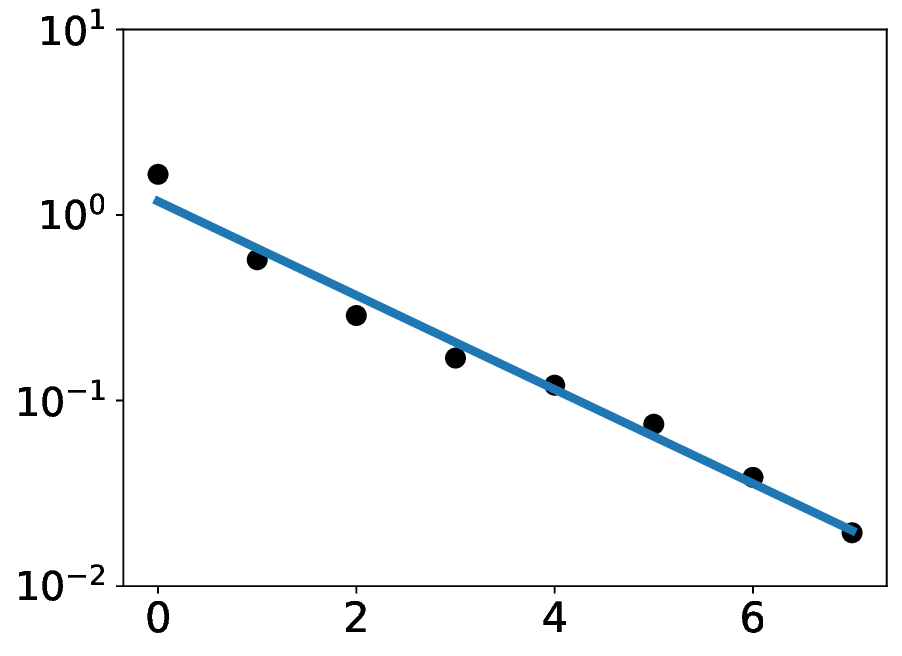}
			\label{fig:class-8}}\hfill
     	\subfloat[Class 9]{
			\centering
			\includegraphics[scale=0.19]{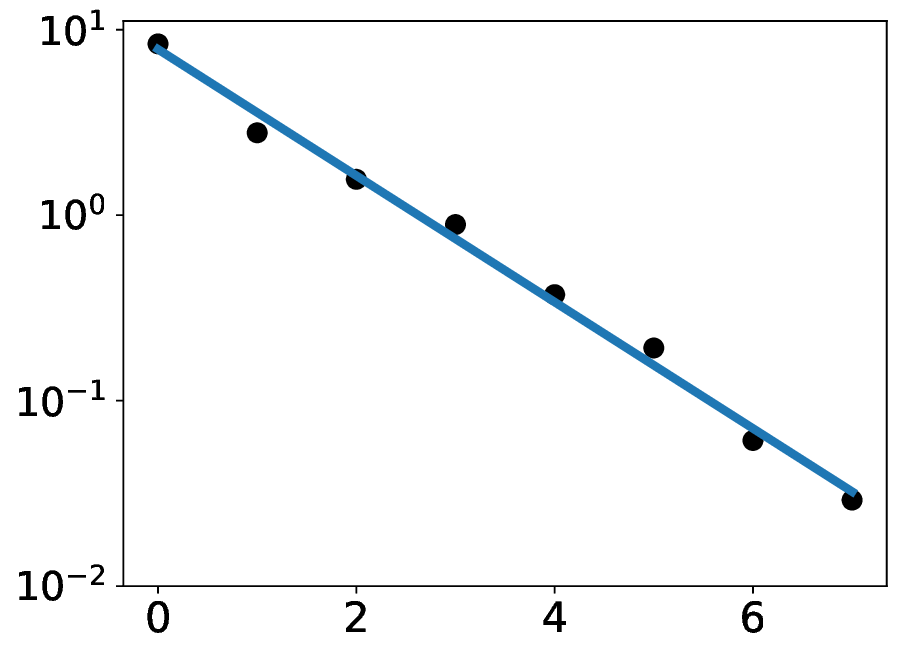}
			\label{fig:class-9}}\hfill
		\subfloat[Class 10]{
			\centering
			\includegraphics[scale=0.19]{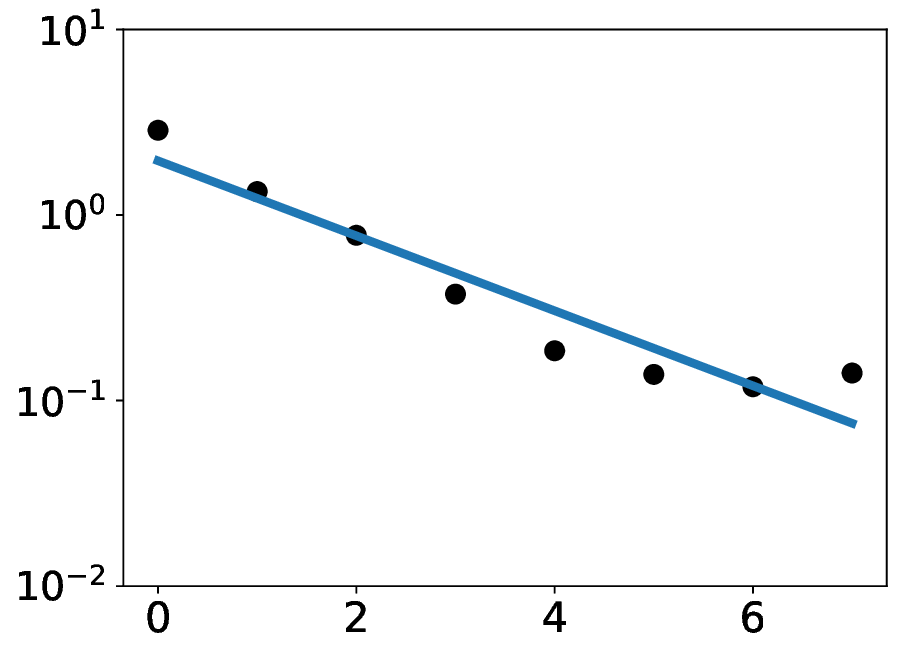}
			\label{fig:class-10}}
			
    \caption{The law of equi-separation emerges for each class.
    }
	\label{fig:class-level}
\end{figure*}

\paragraph{Class-wise data separation.}  As shown in Fig.~\ref{fig:class-level}, we further consider separation fuzziness for each class, i.e., $\ssw^k \ssb^+$, where $\ssw^k$ indicates the within-class covariance for Class $k$.  For simplicity, we only consider the $8$-layer FNNs trained on Fashion-MNIST. The equi-separation law emerges in each class. Even though the equi-separation law can be a little noisy for some classes, the noise is reduced when we consider the separation fuzziness for all classes.

\begin{figure*}[!htp]
	\centering
	\captionsetup[subfigure]{labelformat=empty}
        \subfloat[Layer=0]{
			\centering
			\includegraphics[scale=0.26]{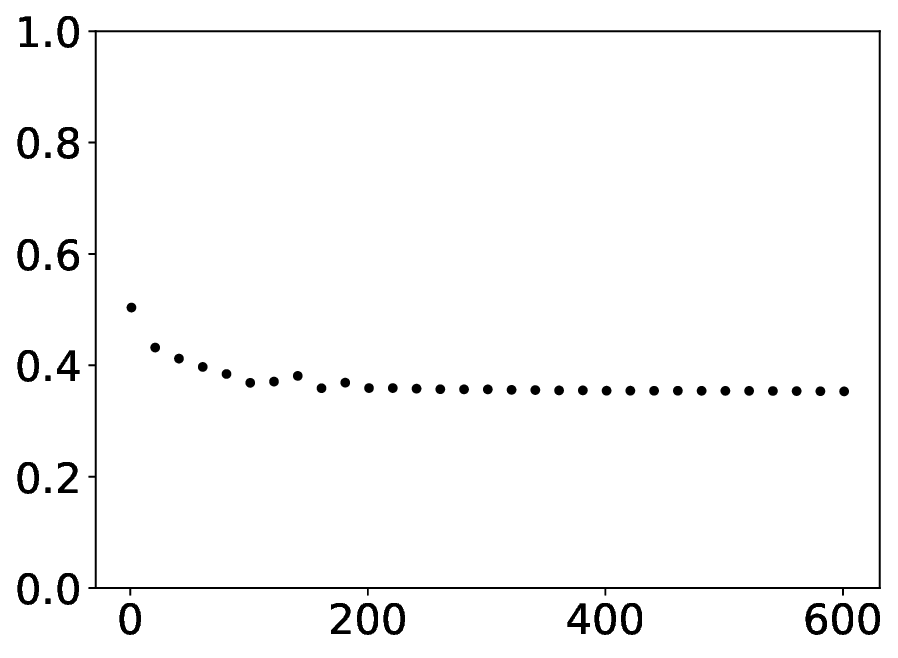}
			\label{fig:ratio-layer-0}}
        	\subfloat[Layer=1]{
			\centering
			\includegraphics[scale=0.26]{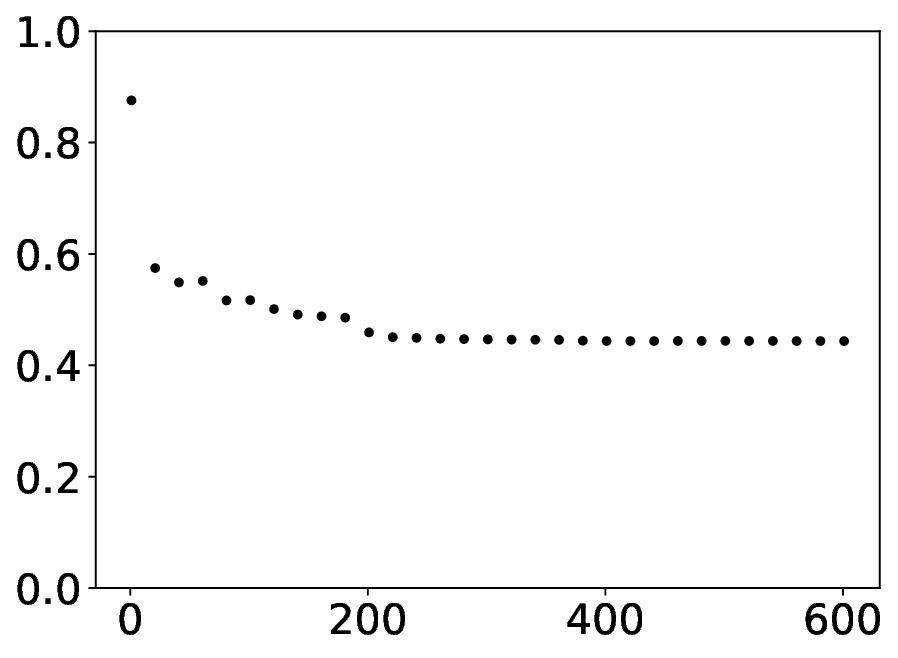}
			\label{fig:ratio-layer-1}}
     	\subfloat[Layer=2]{
			\centering
			\includegraphics[scale=0.26]{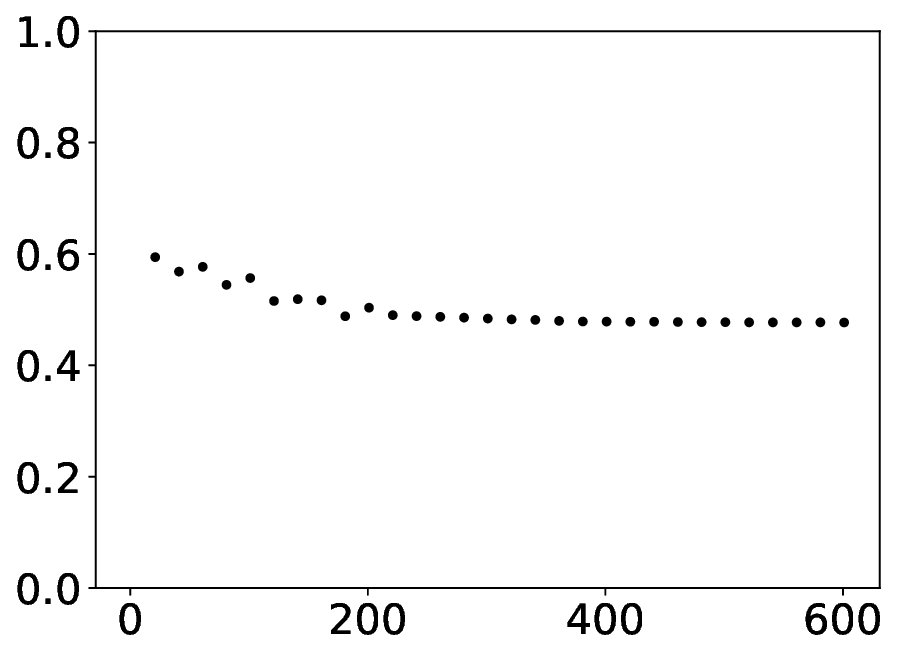}
			\label{fig:ratio-layer-2}}
		\subfloat[Layer=3]{
			\centering
			\includegraphics[scale=0.26]{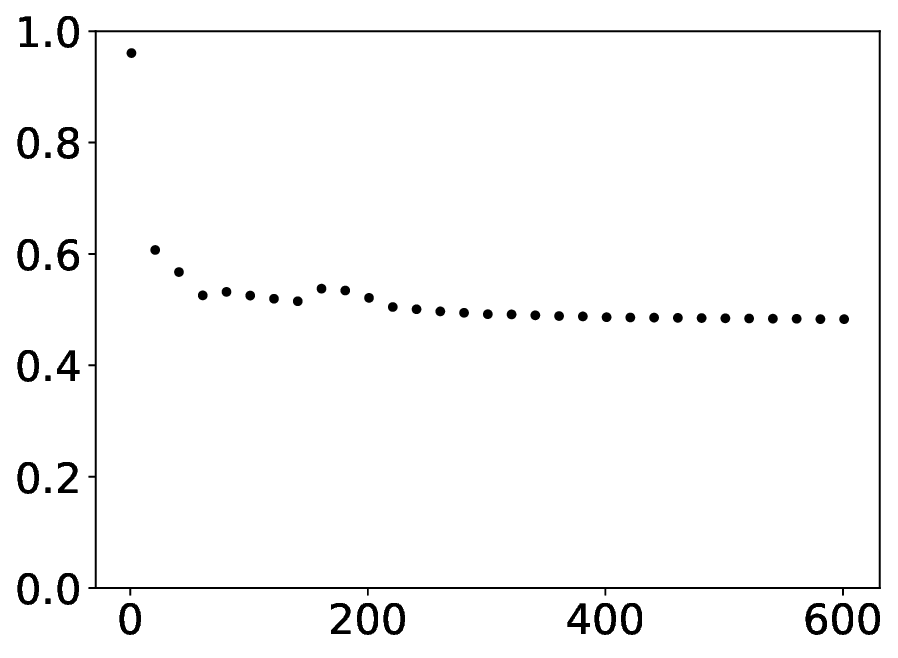}
			\label{fig:ratio-layer-3}}
			
        	\subfloat[Layer=4]{
			\centering
			\includegraphics[scale=0.26]{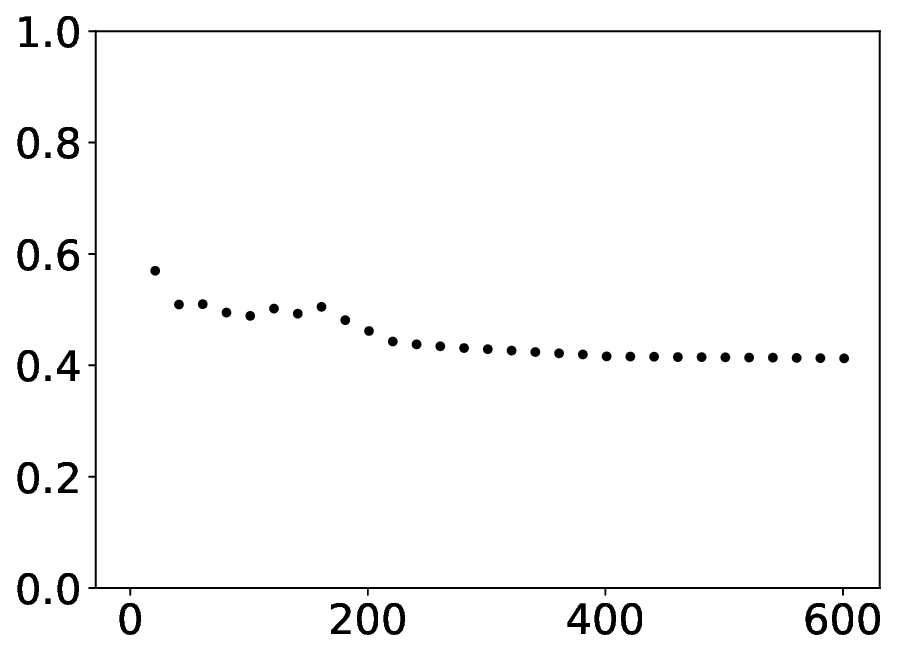}
			\label{fig:ratio-layer-4}}
     	\subfloat[Layer=5]{
			\centering
			\includegraphics[scale=0.26]{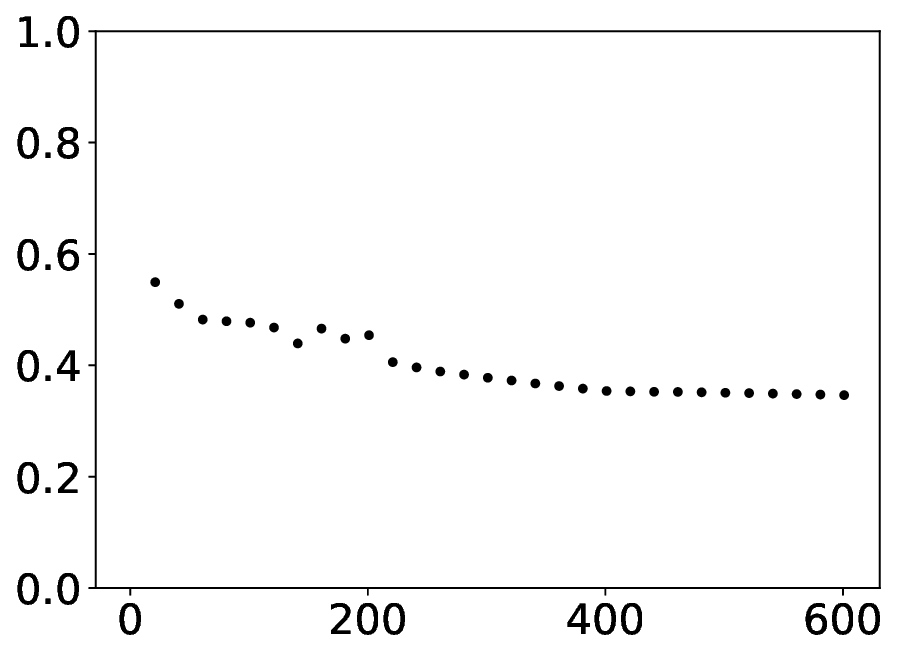}
			\label{fig:ratio-layer-5}}
		\subfloat[Layer=6]{
			\centering
			\includegraphics[scale=0.26]{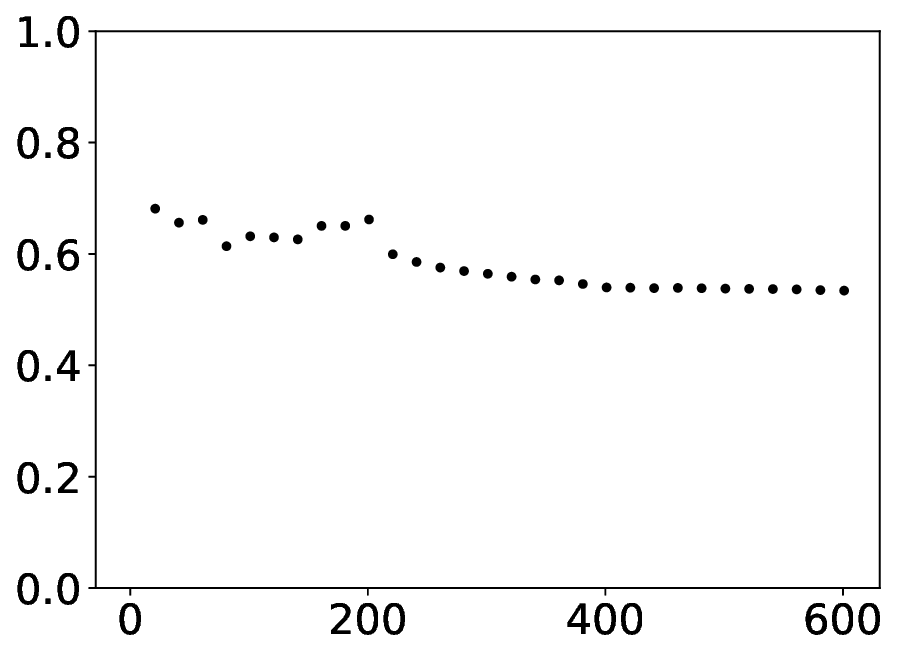}
			\label{fig:ratio-layer-6}}
			
    \caption{Convergence rates of different layers (without the last-layer classifier) with respect to the relative improvement ($\frac{D_{l+1}}{D_l}$) of separability of features. The $x$ axis represents the training epoch, while the y-axis indicates the relative improvement ($\frac{D_{l+1}}{D_l}$).    }
	\label{fig:ratio-dynamic}
\end{figure*}

\paragraph{Training dynamics.} As shown in Fig.~\ref{fig:ratio-dynamic}, we consider the convergence rates of different layers (without the last-layer classifier) with respect to the relative improvement $\frac{D_{l+1}}{D_l}$ of separability of features. For simplicity, we consider the $8$-layer FNNs trained on CIFAR-10 in this part. The relative improvement of bottom layers converges earlier compared to those of top layers.

\begin{figure*}[!htp]
        \captionsetup[subfigure]{labelformat=empty}
		\centering
		 	\subfloat[Depth=4]{
			\centering
			\includegraphics[scale=0.33]{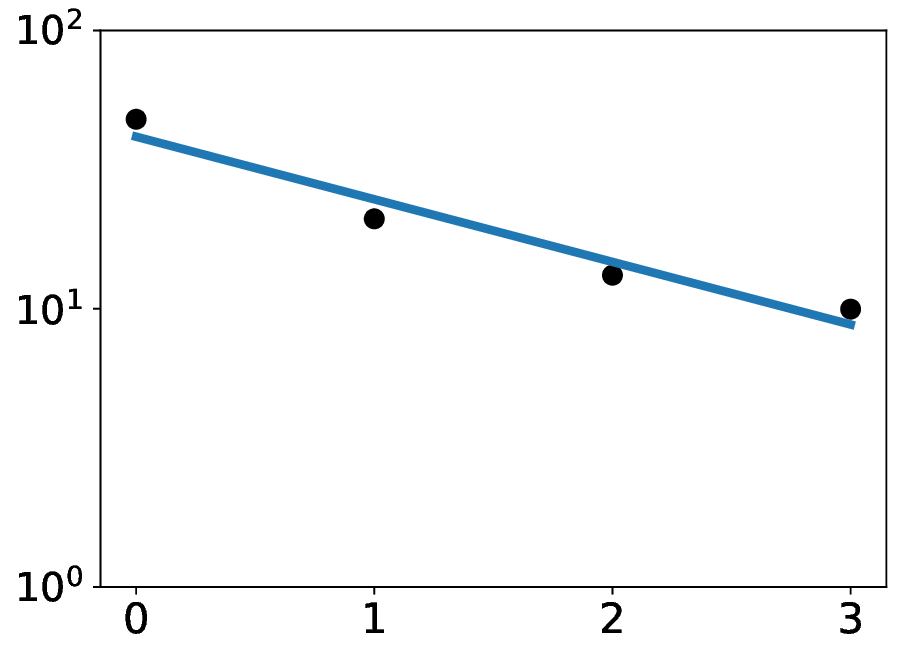}
			\label{fig:adam-4-test}}\hfill
        	\subfloat[Depth=8]{
			\centering
			\includegraphics[scale=0.33]{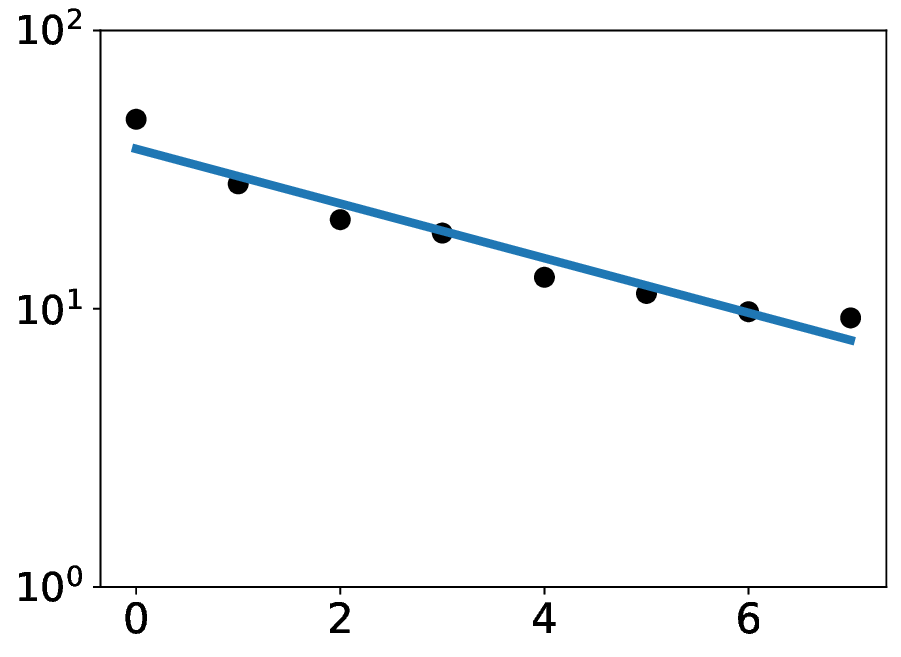}
			\label{fig:adam-8-test}}\hfill
     	\subfloat[Depth=20]{
			\centering
			\includegraphics[scale=0.33]{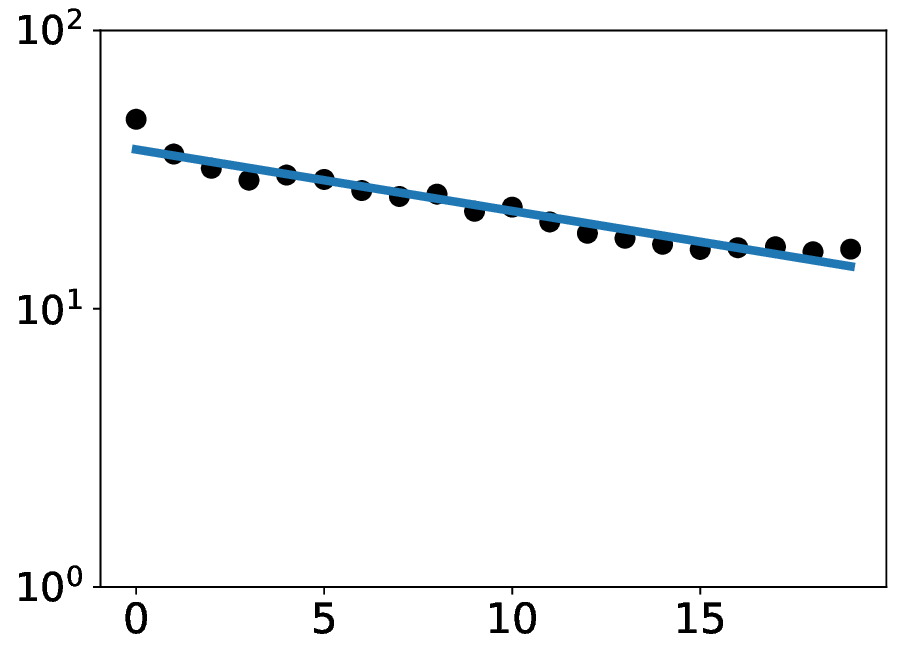}
			\label{fig:adam-20-test}}
		\caption{A fuzzy version of the equi-separation law also emerges in the test data.
		}
		\label{fig:test}
\end{figure*}

\paragraph{Test data.} As shown in  Fig.~\ref{fig:test}, we further show how features separate across layers in the test data. For simplicity, we only consider the $8$-layer FNNs trained on Fashion-MNIST here. A fuzzy version of the equi-separation law also exists in the test data.

\begin{figure*}[!htp]
        \captionsetup[subfigure]{labelformat=empty}
		\centering
		 	\subfloat[BERT-CLS]{
			\centering
    	\includegraphics[scale=0.38]{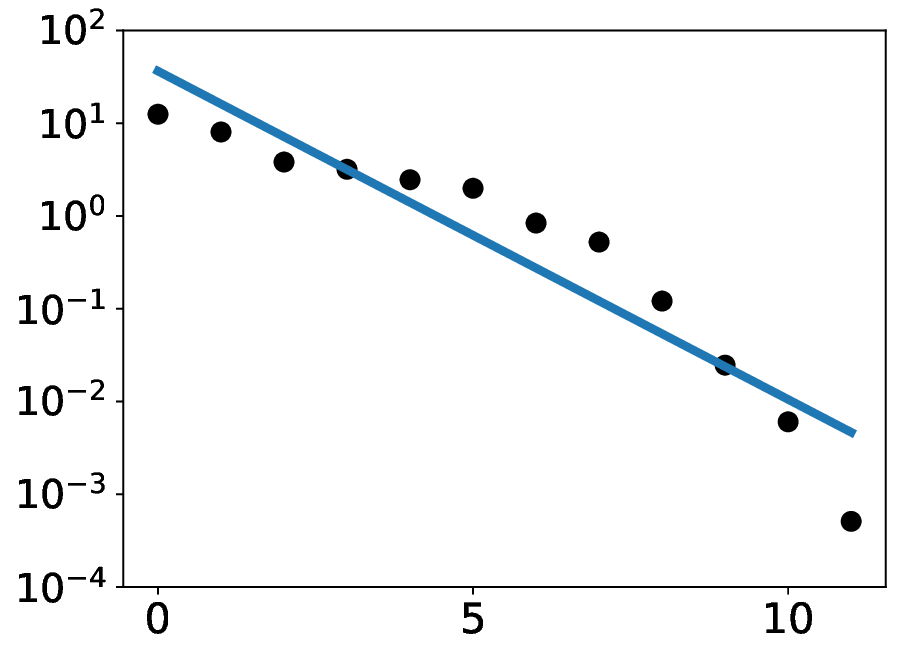}
			\label{fig:bert-cls}}
        	\subfloat[BERT-AVG]{
			\centering
		\includegraphics[scale=0.38]{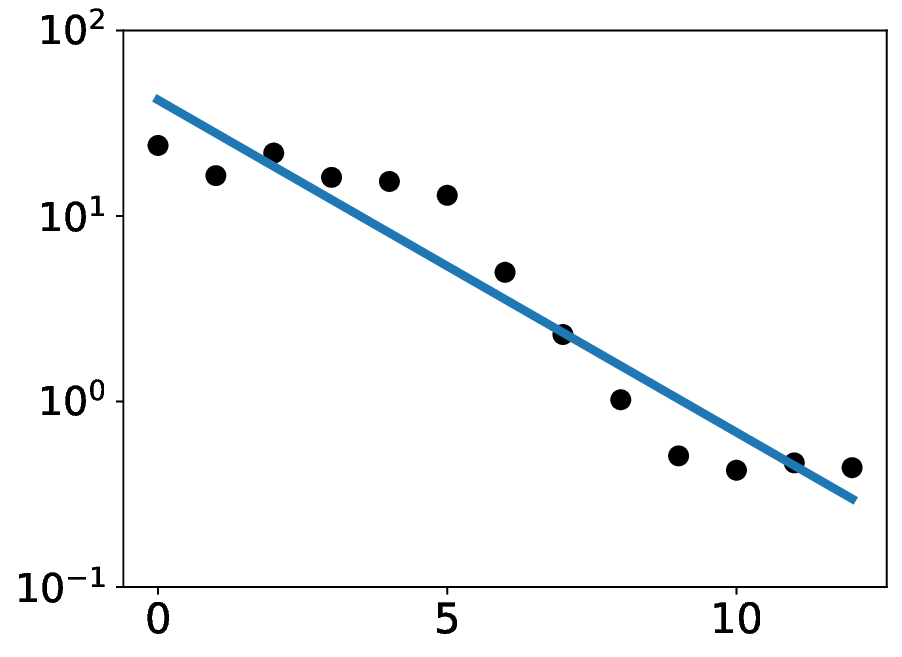}
			\label{fig:bert-avg}}
		
		\caption{The equi-separation law does not exist in BERT features. Given a sequence of token-level BERT features, two most popular approaches are used to get the sentence-level features at each layer: 1) using the features of the first token (i.e., the [CLS] token); 2) averaging the features among all tokens in the sentence (denoted as BERT-AVG).
		}
		\label{fig:bert}
\end{figure*}

\paragraph{BERT.} As shown in Fig.~\ref{fig:bert}, we experiment with BERT features. We use the pretrained case-sensitive BERT-base PyTorch implementation \citep{wolf2020transformers}, and the common hyperparameters for BERT. Specifically, we fine-tuned the pretrained BERT model on a binary sentiment classification task (SST-2) \citep{socher-etal-2013-recursive}, where we randomly sampled $500$ sentences for each class. Given a sequence of token-level BERT features, two most popular approaches are used to get the sentence-level features at each layer: 1) using the features of the first token\footnote{The input layer is not considered here since the inputs of the first token do not take other tokens into account.} (i.e., the [CLS] token); 2) averaging the features among all tokens in the sentence.

\begin{figure*}[!htp]
         \captionsetup[subfigure]{labelformat=empty}
		\centering
		\subfloat[Depth=4]{
			\centering			\includegraphics[scale=0.33]{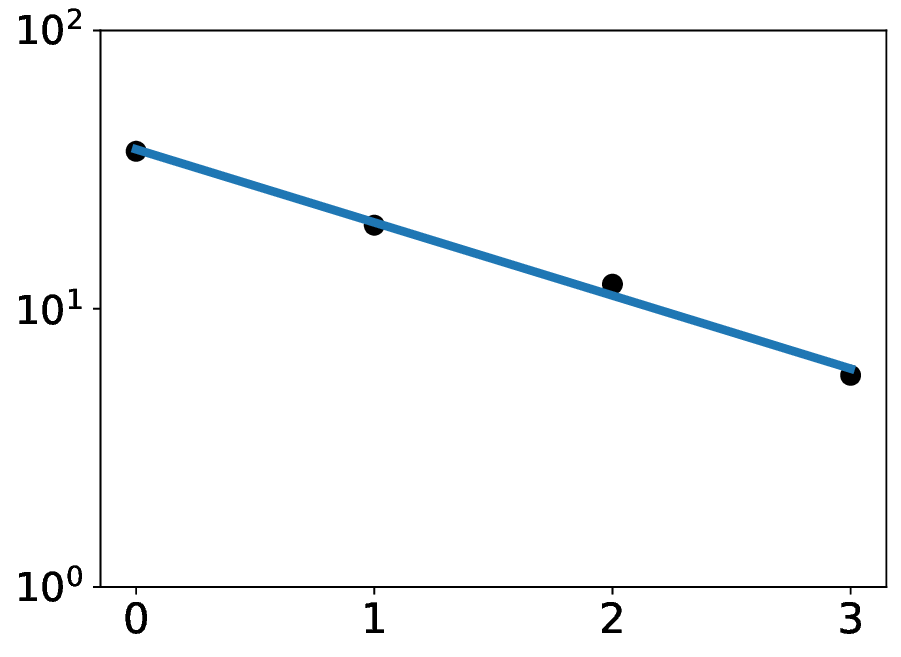}
			\label{fig:NoBN-adam-4}}\hfill
        	\subfloat[Depth=8]{
			\centering			\includegraphics[scale=0.33]{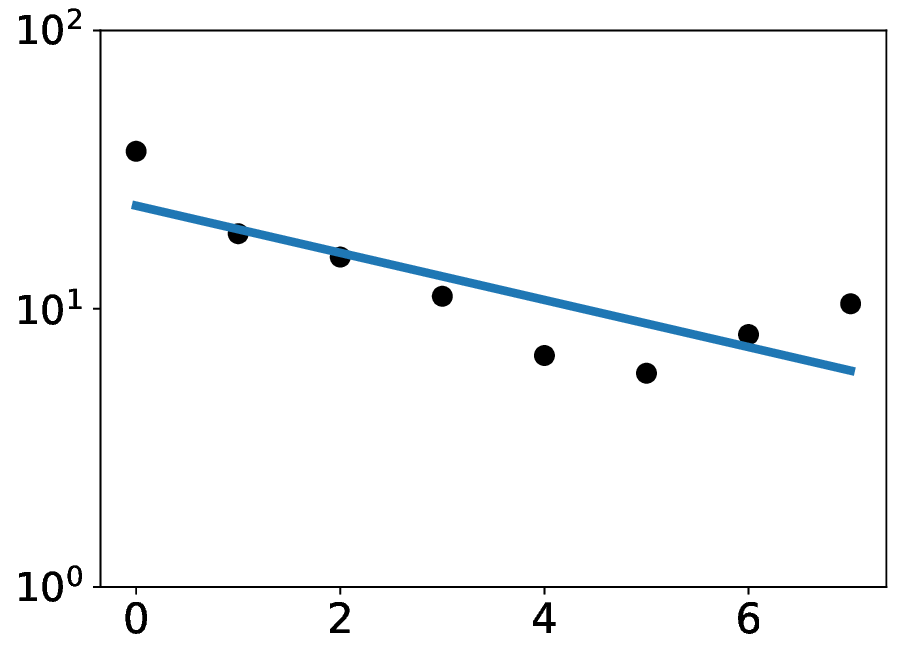}
			\label{fig:NoBN-adam-8}}\hfill
     	\subfloat[Depth=20]{
			\centering	\includegraphics[scale=0.33]{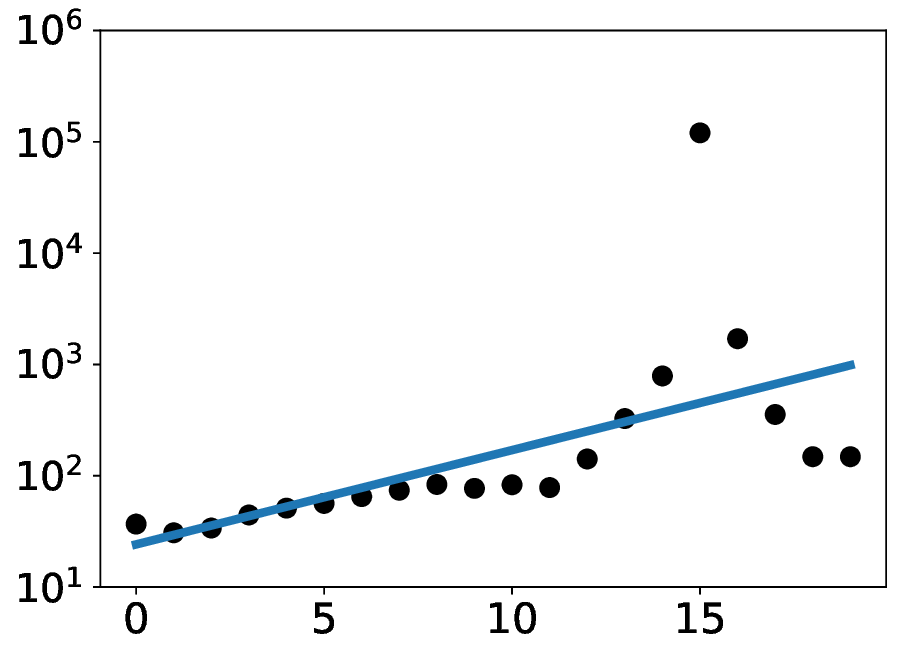}
			\label{fig:NoBN-adam-20}}
			
		\caption{The equi-separation law is not clear on Fashion-MNIST for FNNs trained without batch normalization.}
		\label{fig:NoBN}
\end{figure*}

\paragraph{Batch normalization.} As shown in Fig.~\ref{fig:NoBN}, it is difficult to optimize a neural network without batch normalization, let alone achieve the law of equi-separation. Even when the network is well-optimized, the law is often not clear.

\begin{figure*}[!htp]
        \captionsetup[subfigure]{labelformat=empty}
		\centering
		\subfloat[Depth=20]{
			\centering
			\includegraphics[scale=0.33]{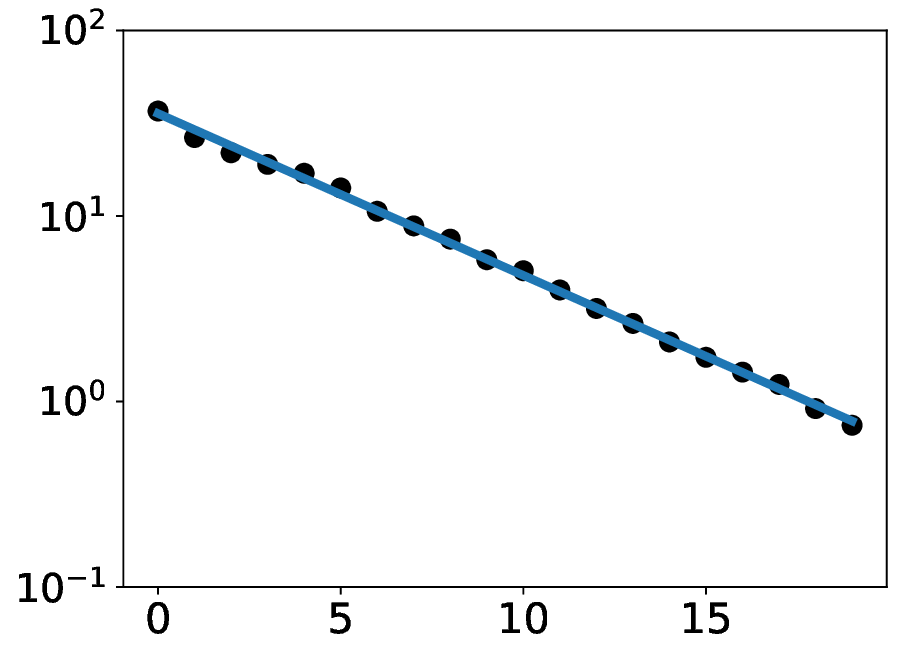}
			\label{fig:layer=20}}\hfill
        	\subfloat[Depth=15]{
			\centering
			\includegraphics[scale=0.33]{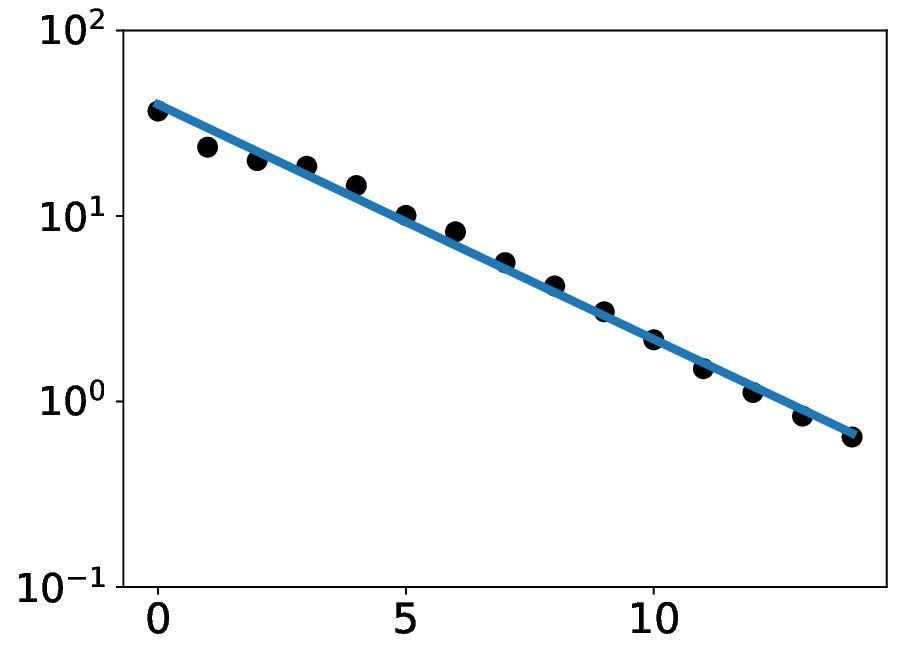}
			\label{fig:layer=15}}\hfill
     	\subfloat[Pretraining]{
			\centering
			\includegraphics[scale=0.33]{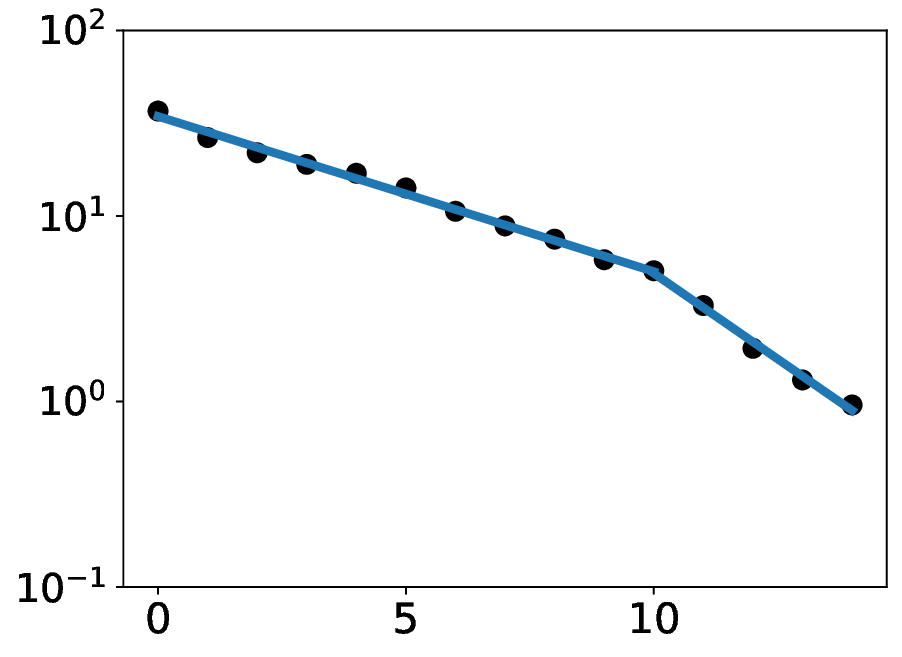}
			\label{fig:pretraining-training}}			
		\caption{The impact of pretraining on the equi-separation law. We first pretrained the $20$-layer FNNs as shown in Fig.~\ref{fig:pretraining} (Depth=20). After that we fix the first $10$ layers of the pretrained model and train additional $5$ layers as in Fig.~\ref{fig:pretraining} (Pretraining), which is quite different from training the $15$-layer neural networks from scratch as in Fig.~\ref{fig:pretraining} (Depth=15).
		}
		\label{fig:pretraining}
\end{figure*}

\paragraph{Pretraining.} As shown in Fig.~\ref{fig:pretraining}, we consider the impact of the pretraining on the equi-separation law. Specifically, we first pretrained the $20$-layer FNNs on FashionMNNIST as shown in Fig.~\ref{fig:pretraining} (Depth=20). After that we fix the first $10$ layers of the pretrained model and train additional $5$ layers as in Fig.~\ref{fig:pretraining} (Pretraining), which is quite different from training $15$-layer FNNs from scratch as in Fig.~\ref{fig:pretraining} (Depth=15). At the same time, the equi-separation law also emerges in the additional $5$ layers in the pretraining setting.

\end{document}